\documentclass[runningheads]{llncs}
\usepackage{graphicx}
\usepackage{amsmath,amssymb} 
\usepackage{color}

\usepackage[width=122mm,left=12mm,paperwidth=146mm,height=193mm,top=12mm,paperheight=217mm]{geometry}

\usepackage{times}
\usepackage{booktabs}
\usepackage{mathrsfs}
\usepackage{algorithm}
\usepackage{algorithmic}
\usepackage{bm}
\usepackage{subfigure}
\usepackage{sidecap}
\usepackage{caption}
\usepackage{multirow}
\usepackage{url}
\usepackage{bbm}

\usepackage[colorlinks, citecolor=blue]{hyperref}
\begin{document}
\title{Adaptively Transforming Graph Matching} 

\titlerunning{Adaptively Transforming Graph Matching}
%
\author{Fudong Wang\inst{1} \and
Nan Xue\inst{1} \and
Yipeng Zhang\inst{2} \and
Xiang Bai\inst{3} \and
Gui-Song Xia\inst{1}\thanks{corresponding author}}

\index{Lastnames, Firstnames}
\authorrunning{F. Wang et al.}
%

\institute{State Key Lab. LIESMARS, Wuhan University, China \\
	\email{\{fudong-wang, xuenan, guisong.xia\}@whu.edu.cn}\\\and
School of Computer Science, Wuhan University, China\\
\email{zyp91@whu.edu.cn}\\
\and
EIS, Huazhong University of Science and Technology, China\\
\email{xbai@hust.edu.cn}}
\maketitle              
\begin{abstract}
Recently,  many graph matching methods that incorporate pairwise constraint and that can be formulated as a quadratic assignment problem (QAP) have been proposed.
Although these methods demonstrate promising results for the graph matching problem, they have high complexity in space or time. 
In this paper, we introduce an adaptively transforming graph matching (ATGM) method from the perspective of functional representation.
More precisely, under a transformation formulation, we aim to match two graphs by minimizing the discrepancy between the original graph and the transformed graph. With a linear representation map of the transformation, the pairwise edge attributes of graphs are explicitly represented by unary node attributes, which enables us to reduce the space and time complexity significantly. Due to an efficient Frank-Wolfe method-based optimization strategy, we can handle graphs with hundreds and thousands of nodes within an acceptable amount of time. 
Meanwhile, because transformation map can preserve graph structures, a domain adaptation-based strategy is proposed to remove the outliers.
The experimental results demonstrate that our proposed method outperforms the state-of-the-art graph matching algorithms.

\keywords{Graph matching  \and Transformation representation \and Frank-Wolfe method}
\end{abstract}
\section{Introduction}\label{introduction}
Graph matching is widely used in a wide range of computer vision and pattern recognition tasks~\cite{[2002-Belongie-pami],[2011-Duchenne],[2012-Yao-eccv],[2016-Shen-eccv],[2016-Garro-pami],[2017-Pinheiro],[Xue2018]} to find correspondence between two graph-structured feature sets.
The general idea behind graph matching solutions is to minimize objective functions composed of unary, pairwise~\cite{[2005-Leordeanu],[2010-Cho-eccv],[2017-Huu-cvpr]} or higher-order~\cite{[2011-Lee-cvpr],[2015-Yan-cvpr],[2017-Huu-cvpr],[YU2016255]} potentials to preserve the structure alignment between two graphs.

\begin{figure}[htb!]
	\centering
	\includegraphics[width=0.9\linewidth]{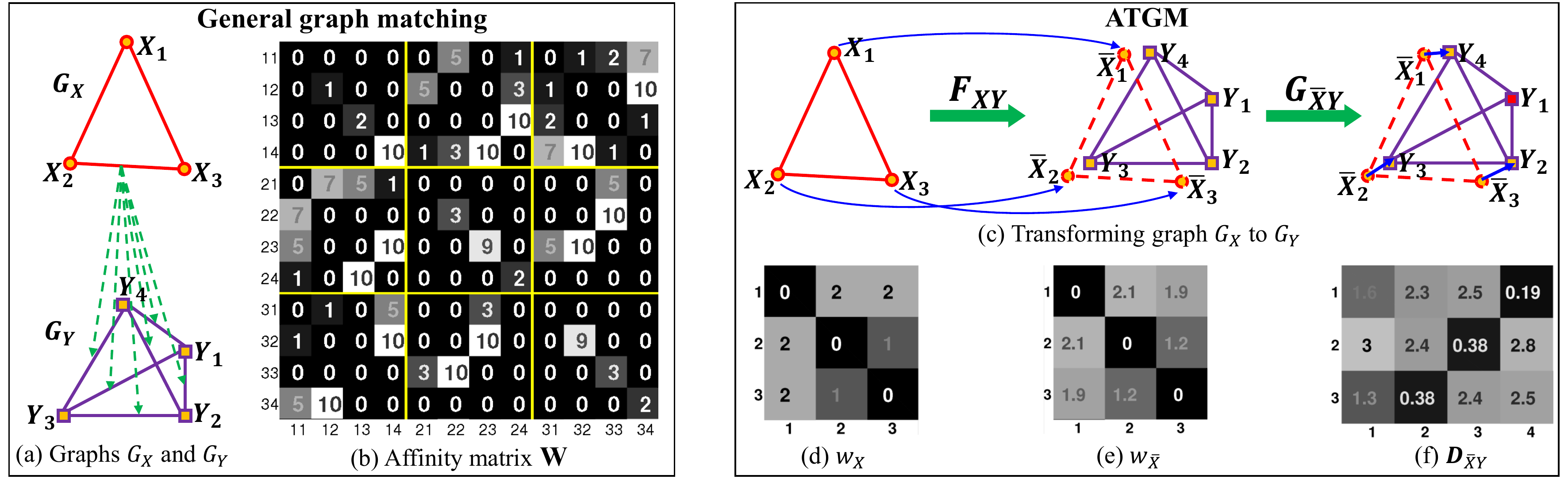}
	\caption{{\bf ATGM vs. general graph matching}. General graph matching between two graphs $\mathcal{G}_X$ and $\mathcal{G}_Y$ with nodes $X =\{X_i\}_{i=1}^3$ and $Y =\{Y_j\}_{j=1}^4$, shown in (a), often involves computing a pairwise affinity matrix $\mathbf{W}$ with a size of $12\times 12$, as displayed in (b). In contrast, our proposed ATGM matches $\mathcal{G}_X$ to $\mathcal{G}_Y$ by transforming each node $X_i$ to $\bar{X}_i$ with an optimal transformation map through minimizing two objectives $F_{XY}$ and $G_{\bar{X}Y}$, as shown in (c). ATGM preserves the pairwise structure alignments by minimizing the differences between two edge attribute matrices $w_{X}$ and $w_{\bar{X}}$ with a smaller size of $3\times 3$, as shown in (d) and (e). ATGM can also remove the outliers, {\it i.e.}, $Y_1$ here, naturally according to the distance matrix $\mathbf{D}_{\bar{X}Y}$ between $\mathcal{G}_{\bar{X}}$ and $\mathcal{G}_{Y}$, as shown in (f).}
	\label{fig:1}
\end{figure}

Under pairwise constraint, graph matching can be formulated as a quadratic assignment problem (QAP)~\cite{[2007-Loiola-ejor]}, which is NP-complete~\cite{[1979-Garey]}, and only approximate solutions can be found in polynomial time.
Although the past decade has witnessed remarkable progress on graph matching~\cite{[2016-Yan-ICMR],[2009-Leordeanu-nips],[2010-Cho-eccv],[2014-Cho-cvpr],[2016-Zhou-pami]}, there are still some challenges in computational complexity and matching performance.
For instance, a costly affinity matrix often needs to be computed or factorized~\cite{[2009-Leordeanu-nips],[2005-Leordeanu],[2016-Zhou-pami]}, which results in high space complexity--especially with large-scale complete graphs. 
Because of the combinatorial nature of QAP, the objective function is difficult to solve for obtaining binary solutions~\cite{[2009-Leordeanu-nips],[2015-Yan-cvpr],[2010-Lee-icpr]}.
Although with relaxation, the discrete constraint can be approximated by a continuous one that is easier to solve, this approach requires extra effort to achieve a global optimum or satisfy the binary constraint~\cite{[2009-Zaslavskiy-pami],[2016-Zhou-pami],[2014-Liu-pami],[2017-Jiang-cvpr]}.
Moreover, matching unequal-sized graphs often suffer from outliers~\cite{[2014-Cho-cvpr],[2009-Zaslavskiy-pami]}. Thus, it is of great interest to reduce the computational complexity and to be as robust as possible to outliers.

This paper introduces a method for graph matching from the perspective of functional representation.
The main idea is illustrated by a toy example in Fig.~\ref{fig:1}. 
Under this perspective, one graph is transformed into the space spanned by the second graph, and then, a desired correspondence can be reformulated as an optimal transformation map between graphs. 
To pursue such a  map, we construct two functionals to measure the discrepancy between graphs and minimize them with the Frank-Wolfe method~\cite{[2015-Simon-nips]}.
Using the transformation map, the pairwise edge attributes of graphs can be explicitly represented by node attributes, which enables us to significantly reduce the space and time complexity. We also propose a domain adaptation-based strategy to remove outliers leveraging the fact that transformation maps can preserve graph structures. 

Our work is distinguished in following aspects:
\begin{itemize}
	\item[-] We present a new perspective for graph matching that explicitly represents the pairwise edge attributes of graphs using unary node attributes. Therefore,
	the space complexity is reduced in form from $\mathbf{O}(m^2n^2)$ to $\mathbf{O}(mn)$ and
	the objective function can be optimized efficiently with $\mathbf{O}(Tn^3)$ time complexity. Benefiting from this simplification, we can match large-scale graphs, even with complete graphs.
	
	\item[-] We propose a domain adaptation-based method for outlier removal using the transformation map. This technique can be used as a pre-processing step to improve graph matching algorithms.
\end{itemize}

\section{Related Work}\label{relatedwork}

Over the past few decades, both exact and inexact (error-tolerant) graph matching have been extensively studied to measure either (dis-)similarity~\cite{[2011-Duchenne],[1999-Pelillo-pami],[2009-Riesen-pami],[2017-Bougleux]} or find correspondence~\cite{[1996-Gold],[2009-Leordeanu-nips],[2010-Cho-eccv],[2015-Yan-cvpr],[2016-Zhou-pami]} between graphs. We focus on inexact graph matching to find correspondence as the work on exact graph matching and measuring similarity ({\itshape e.g.}, graph edit distance) is beyond the scope of this paper. 

Many existing works of pairwise graph matching have addressed reducing the high computational complexity of the QAP formulation. In the path-following method proposed in~\cite{[2009-Zaslavskiy-pami]}, the author rewrote the graph matching problem as an approximate least-squares problem on the set of permutation matrices. A factorization-based method~\cite{[2016-Zhou-pami]} was proposed to factorize the affinity matrix with high space complexity into a Kronecker product of smaller matrices. An efficient sampling heuristic has been proposed in~\cite{[2008-Zass-cvpr]} to avoid the high space complexity of the affinity matrix. However, the methods in~\cite{[2009-Zaslavskiy-pami],[2016-Zhou-pami]} suffer from huge time consumption in practice, and the ability to reduce space complexity of the works~\cite{[2016-Zhou-pami],[2008-Zass-cvpr]} is limited by complete graphs. As a comparison, our functional representation-based method can reduce the space complexity by two orders of magnitude with a lower time complexity and runs faster in practice.

Considering looking for global optimal solutions with binary property for graph matching, the approaches in~\cite{[2009-Zaslavskiy-pami],[2016-Zhou-pami],[2014-Liu-pami],[2014-Liu-ijcv]} constructed objective functions in both convex and concave relaxations that were controlled by a continuation parameter. However, these approaches are often time consuming in reaching an ideal solution. Moreover, to ensure binary solutions, an integer-projected fixed point algorithm~\cite{[2009-Leordeanu-nips]} solving a sequence of first-order Taylor approximations had been proposed, and the author of~\cite{[2015-Yan-cvpr]} took an adaptive and dynamic relaxation mechanism for optimization in the discrete domain directly. In our method, we separately construct non-convex and convex relaxations and obtain (nearly) binary solutions in a faster way with high matching accuracy.

In addition, several spectral matching methods~\cite{[2005-Leordeanu],[2006-Cour-nips]} were introduced based on the rank-1 approximation of the affinity matrix. The graduated assignment method~\cite{[1996-Gold]} iteratively solved a series of convex approximations of the objective. The decomposition-based works in~\cite{[2013-Torresani-pami]} and~\cite{[2017-Huu-cvpr]} decomposed the original complex graphs and took decomposition of the matching constraints,respectively. Probability-based~\cite{[2008-Zass-cvpr],[2013-Egozi]} and learning-based~\cite{[2009-Caetano-pami],[2012-Leordeanu-ijcv]} methods gave further interpretations of the graph matching problem.  A random walk view~\cite{[2010-Cho-eccv]} of the problem was introduced by simulating random walks with re-weighting jumps. A max-pooling based strategy has been also proposed in~\cite{[2014-Cho-cvpr]} to address the presence of outliers. These two works~\cite{[2010-Cho-eccv],[2014-Cho-cvpr]} are both robust to outliers due to their re-weighting procedure during iterations. In contrast, our proposed outlier-removal strategy removes the outliers by explicitly relying on the global structure of graphs, and it can be applied to other methods as a pre-processing step.

\section{General Graph Matching}

Given an undirected graph $\mathcal{G}_X=\left\{X,\mathcal{E}_X\right\}$
with $m$ nodes $X_i\in X,\, i=1,\ldots, m$, we denote each edge as $X_{i_1i_2}\triangleq(X_{i_1},X_{i_2})\in \mathcal{E}_X$, where
$\mathcal{E}_X$ is the edge set consisting of $M$ edges.
Matching the two graphs $\mathcal{G}_X$ and $\mathcal{G}_Y$, with $m,n$ nodes and $M,N$ edges, respectively, yields a binary correspondence $\mathbf{P}\in \left\{0,1\right\}^{m\times n}$,
such that $P_{ij}=1$ when the nodes $X_i$ and $Y_j$ are matched and $P_{ij}=0$ otherwise.

The graph matching problem is often solved by maximizing an objective function that measures the node and edge affinities between $\mathcal{G}_X$ and $\mathcal{G}_Y$.
Under pairwise constraints, the objective function typically consists of a unary potential $w_v(X_i,Y_j)$ and a pairwise potential $w_e(X_{i_1i_2},Y_{j_1j_2})$, which measure the similarity between the nodes $X_i$ and $Y_j$ and the edges $X_{i_1i_2}$ and $Y_{j_1j_2}$, respectively.
These two types of similarities are usually integrated by an affinity matrix $\mathbf{W}\in \mathbb{R}^{mn\times mn}$, the diagonal element $\mathbf{W}_{ij,ij}$ of which corresponds to the unary potential $w_v(X_i,Y_j)$ and the non-diagonal element $\mathbf{W}_{i_1j_1,i_2j_2}$ of which corresponds to the pairwise potential $w_e(X_{i_1i_2},Y_{j_1j_2})$.
Thus, the objective function for graph matching can be written as
\begin{align}
\mathbf{P}_v^T\mathbf{W}\mathbf{P}_v = \sum_{P_{ij}=1}w_v(X_i,X_j) + \sum_{{\tiny{\substack{P_{i_1j_1}=1\\P_{i_2j_2}=1}}}}w_e(X_{i_1i_2},Y_{j_1j_2}),
\label{eq:gm=general}
\end{align}
where $\mathbf{P}_v$ is the column-wise vectorized replica of $\mathbf{P}$.

For graph matching under one-to-(at most)-one
constraints, the feasible field $\mathcal{P}$ is composed of all (partial) permutation matrices (where $m\le n$), {\it i.e.}
\begin{equation}\label{constraint1}
\mathcal{P}\triangleq \left\{\mathbf{P}\in \left\{0,1\right\}^{m\times n}; \mathbf{PI}_n=\mathbf{I}_m,\mathbf{P}^T\mathbf{I}_m\le \mathbf{I}_n\right\},
\end{equation}
where $\mathbf{I}_m$ is a $m \times 1$ unit vector.
Then, the graph matching problem can be approached by finding the optimal assignment matrix $\mathbf{P}^*$ by maximizing
\begin{equation}\label{QAP}
\max_{\mathbf{P}\in \mathcal{P}}~\mathbf{P}_v^T\mathbf{W}\mathbf{P}_v.
\end{equation}

Eq.(\ref{QAP}) is the so-called (QAP), which is known to be NP-complete. Usually, an approximate solution of it can be found by relaxing the discrete feasible field $\mathcal{P}$ into a continuous feasible filed $\hat{\mathcal{P}}$ as:
\begin{equation}\label{constraint2}
\hat{\mathcal{P}}\triangleq \left\{\mathbf{P}\in [0,1]^{m\times n}; \mathbf{PI}_n=\mathbf{I}_m,\mathbf{P}^T\mathbf{I}_m\le \mathbf{I}_n\right\},
\end{equation}
which is known as the {\em doubly-stochastic relaxation}.
Unfortunately, (1) the affinity matrix $\mathbf{W}$ results in high space complexity--especially with complete graphs, and (2) achieving global optimal or binary solutions of Eq.(\ref{QAP}) is often highly time consuming.


\section{Adaptively transforming graph matching}\label{method}

This section presents our ATGM algorithm starting with a definition of the linear representation map of transformation from one graph to the space spanned by another graph. Basically, the transformation map models the correspondence between graphs. On this basis, we first measure the edge discrepancy between two graphs to derive the sub-optimal transformation map. Then, we incorporate the shifting vectors of the transformed nodes to obtain the final optimal transformation map. Finally, we address the unequal size cases in graph matching by proposing a domain adaptation-based outlier removal strategy.

\subsection{Linear representation of transformation}
Given two undirected graphs $\mathcal{G}_X=\left\{X,\mathcal{E}_X \right\}$ and $\mathcal{G}_Y=\left\{Y,\mathcal{E}_Y \right\}$, we formulate graph matching as transformation from node set $X=\left\{X_i \right\}_{i=1}^m$ to the space spanned by $Y=\left\{Y_j \right\}_{j=1}^n$. Because $X,Y$ are discrete sets, we first define the continuous space spanned by $Y$ as $\mathcal{C}_Y={\sum_{i=1}^n \alpha_jY_j}$. Transformation $\mathcal{T}$ from $X$ to $\mathcal{C}_Y$ is defined as
\begin{align}
\mathcal{T}: X \to \mathcal{C}_Y, X_i \mapsto \mathcal{T}(X_i).
\end{align}

According to linear algebra, $\mathcal{T}(X_i)$ can be represented as $\mathcal{T}(X_i)\triangleq\sum_{j=1}^{n}\mathbf{P}_{ij}Y_j$. Then, $\mathbf{P}\in \mathbb{R}^{m\times n}$ is a linear representation ({\itshape i.e.}, a transformation map) of $\mathcal{T}$. By the constraint Eq.(\ref{constraint2}) that $\mathbf{P}\in \hat{\mathcal{P}}$, each node $\mathcal{T}(X_i)$ lies in the convex hull of $Y$. Therefore, we redefine $\mathcal{C}_Y$ as the convex hull of $Y$ for graph matching problem. Whenever $\mathbf{P}$ reaches an extreme point of the feasible field $\hat{\mathcal{P}}$, it is a binary assignment matrix, and consequently, $X_i$ is transformed to ({\itshape i.e.}, matches) a $Y_{j'}$ where $\mathbf{P}_{ij'} =1,\mathbf{P}_{i,j\neq j'}=0$.

By this representation formulation, the transformed graph $\bar{X}\triangleq\mathcal{T}(X)=\mathbf{P}Y$ is determined by specified $\mathbf{P}$ and $Y$. The more $\mathbf{P}$ is binary, the more $\bar{X}$ is similar to $Y$. Therefore, we can replace $\mathcal{G}_Y$ by $\mathcal{G}_{\bar{X}}$ when we attempt to minimize the disagreement between $\mathcal{G}_X$ and $\mathcal{G}_Y$ by forcing $\mathbf{P}$ to be binary. 
With notation $\bar{X}_{i_1i_2}\triangleq(\bar{X}_{i_1},\bar{X}_{i_2})$, we construct the functional w.r.t. $\mathbf{P}$ to measure disagreement between $\mathcal{G}_X$ and $\mathcal{G}_{\bar{X}}$ as
\begin{equation}
\mathbb{F}(\mathbf{P})=\sum_{(i,j)}f_v(X_i,Y_j)P_{ij} + \sum_{(i1,i_2)} f_e(X_{i_1i_2},\bar{X}_{i_1i_2}),
\end{equation}
where the unary potential $f_v(X_i,Y_j)$ denotes the disagreement between nodes $X_i$ and $Y_j$, and the pairwise potential $f_e(X_{i_1i_2},\bar{X}_{i_1i_2})$ denotes the discrepancy between
edge $X_{i_1i_2}$ and its transformed edge $\bar{X}_{i_1i_2}$. Using this formulation, the costly affinity matrix $\mathbf{W}\in \mathbb{R}^{mn\times mn}$ used in general graph matching is replaced by the node disagreement matrix $\left\{ f_v(X_i,Y_j) \right\}\in \mathbb{R}^{m\times n}$ and the edge discrepancy matrix $\left\{ f_e(X_{i_1i_2},\bar{X}_{i_1i_2})\right\}\in \mathbb{R}^{m\times m}$, which consequently reduces the space complexity from $O(m^2 n^2)$ to $O(mn)$.

To obtain a desired assignment matrix $\mathbf{P}^*$ given graphs $\mathcal{G}_X$ and $\mathcal{G}_Y$, we can construct a specified functional $\mathbb{F}(\mathbf{P})$ and minimize it to preserve the structure alignments between $\mathcal{G}_X$ and $\mathcal{G}_{\bar{X}}$ in a optimization-based way:
\begin{equation}
\mathbf{P}^*\in \text{arg}\min\limits_{{\small{\mathbf{P}\in \hat{\mathcal{P}}}}} \mathbb{F}(\mathbf{P}).
\end{equation}

In the rest of this section, we introduce two functionals w.r.t. $\mathbf{P}$ as our objective functions to model the pairwise graph matching problem.

\begin{figure}[htb!]
	\begin{center}
		\subfigure[]
		{\includegraphics[width=0.36\linewidth]{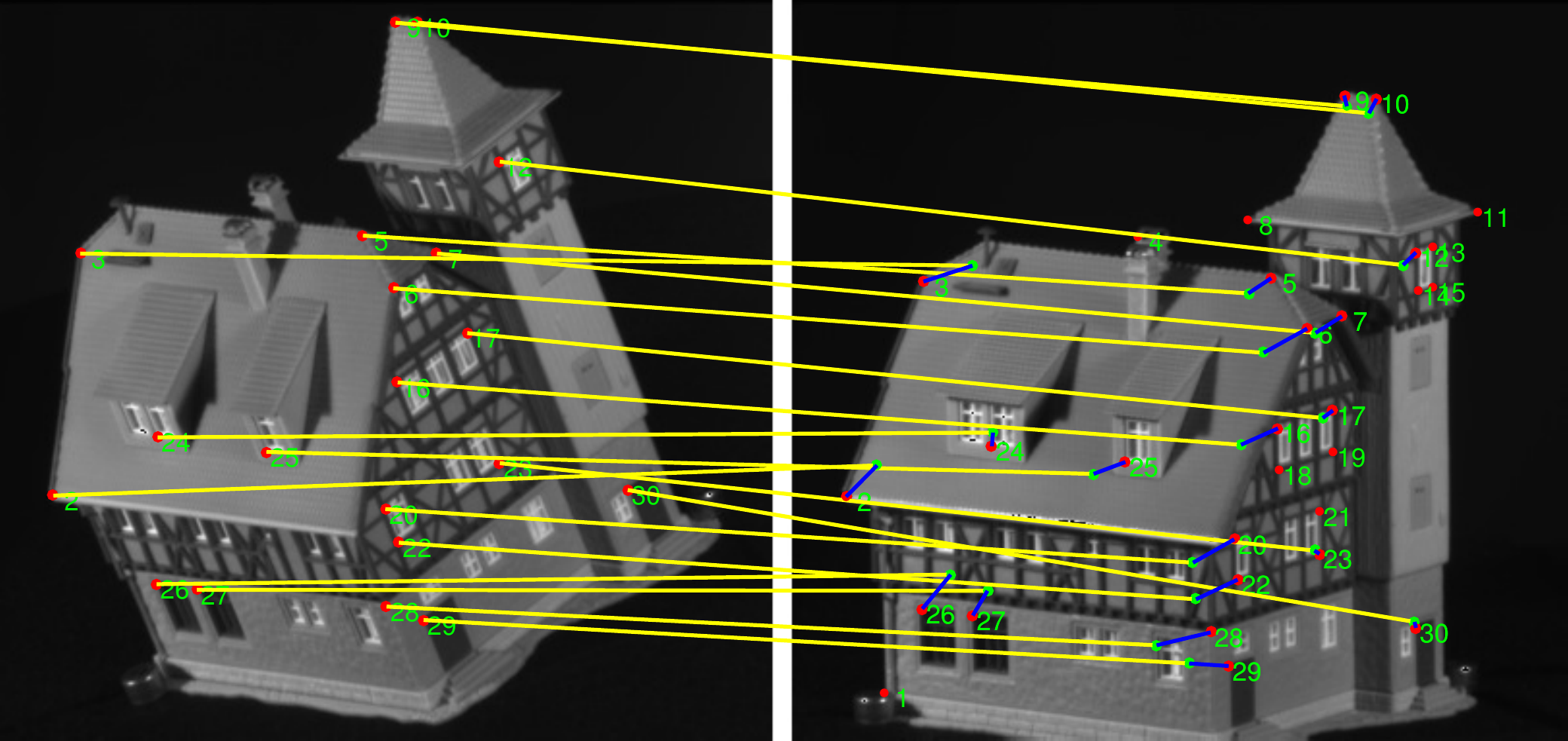}}
		\subfigure[]
		{\includegraphics[width=0.11\linewidth]{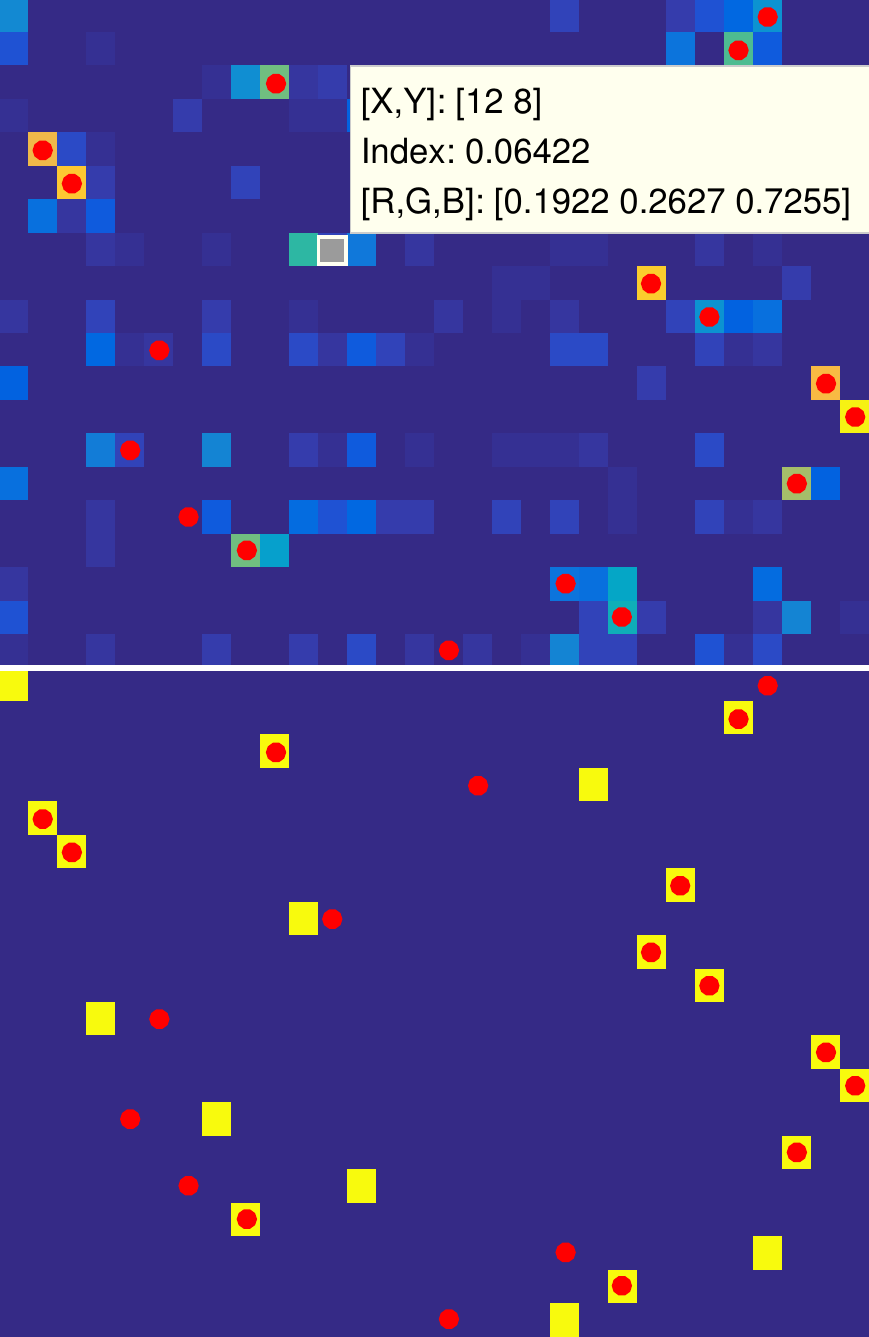}}
		\subfigure[]
		{\includegraphics[width=0.36\linewidth]{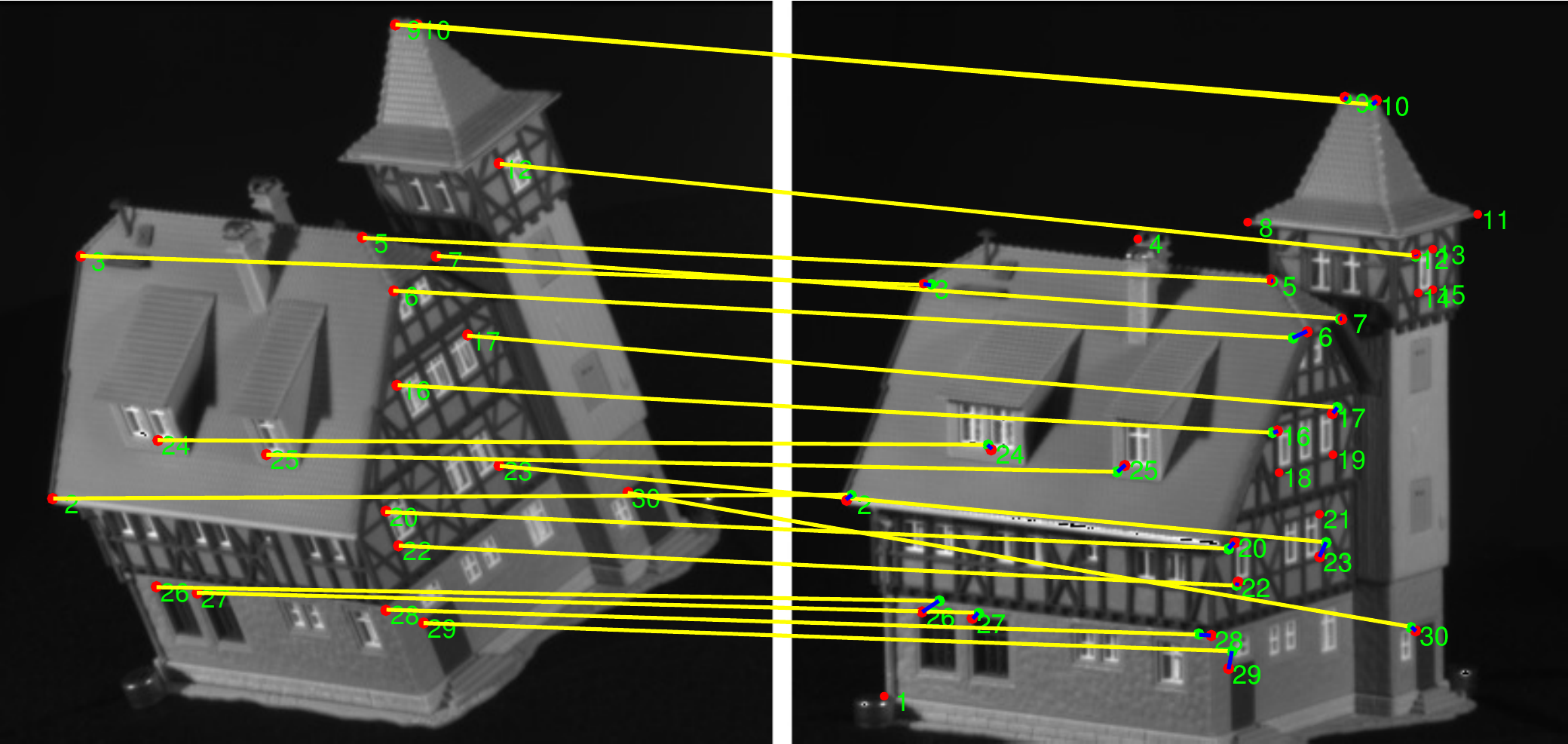}}
		\subfigure[]
		{\includegraphics[width=0.11\linewidth]{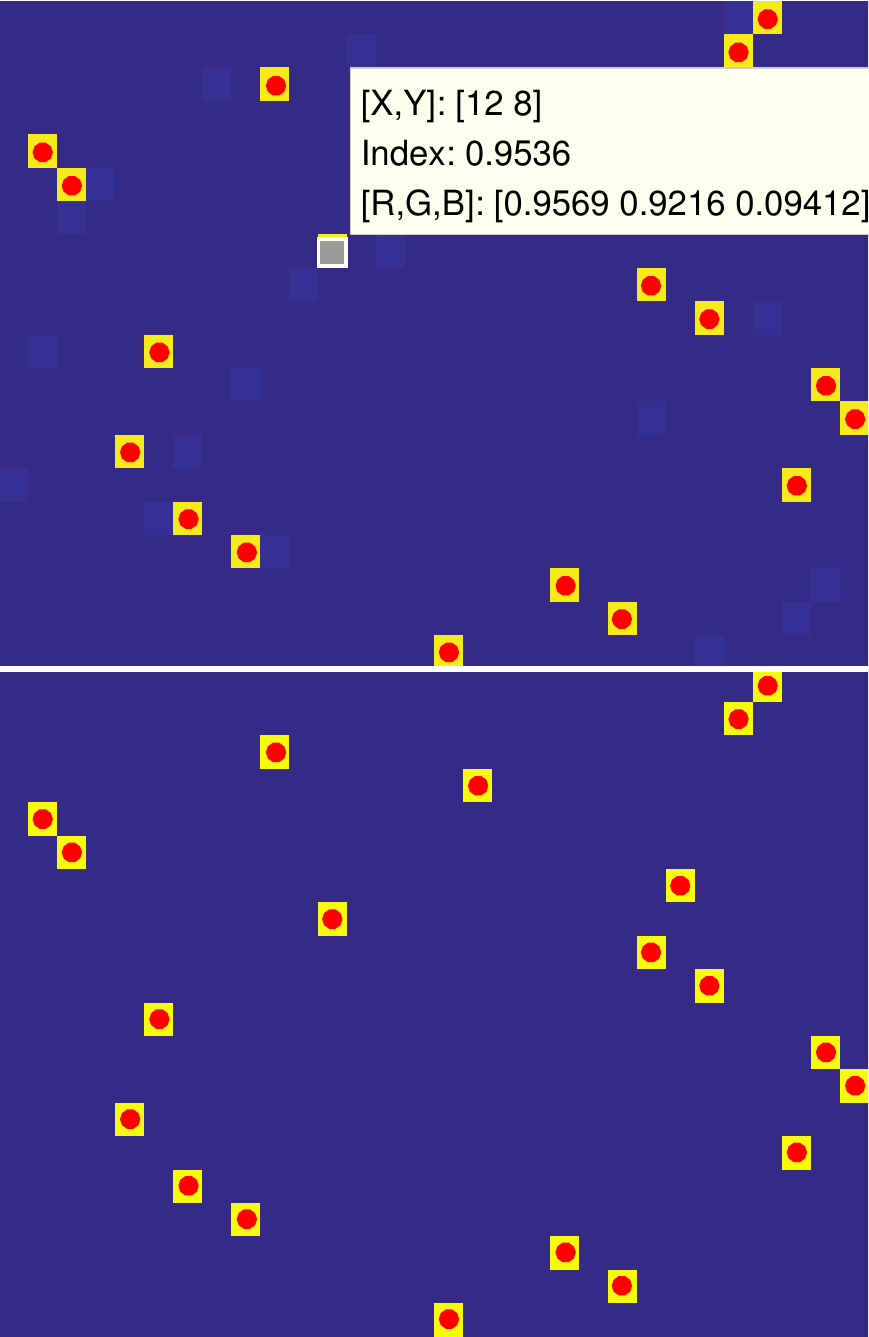}}
	\end{center}
	\caption{(a) Nodes shift after being transformed by minimizing $F_{XY}(\mathbf{P})$ in a 20-vs-30 case. The lines in blue are the shifting vectors, and the points in green are transformed nodes $\left\{\bar{X}_i\right\}_{i=1}^m$. 
		(b) Transformation map (top) and their post-discretization (bottom) corresponding to (a). (c) Nodes transformed by minimizing $G_{\bar{X}Y}(\mathbf{P})$ with almost no shifting. 
		(d) Transformation map (top) and their post-discretization (bottom) corresponding to (c). In (b) and (d), red points mark the groundtruth.}
	\label{fig:long}
	\label{fig:longfig1}
\end{figure}

\subsection{Edge discrepancy}
In the case where graphs are embedded in Euclidean space $\mathbb{R}^d$, the function $f_e$ mentioned above can be defined in some simple but effective forms to incorporate the edge length (or orientations),
\begin{equation}
f_e(X_{i_1i_2},\bar{X}_{i_1i_2}) = (||X_{i_1i_2}||-||\bar{X}_{i_1i_2}||)^2,
\end{equation}
where $||X_{i_1i_2}||$ is the $l_2$ norm of $X_{i_1i_2}$.

Thus, the pairwise potential of our first objective function is defined as,
\begin{align}
F_{XY}(\mathbf{P})&=\sum_{(i_1,i_2)}S_{i_1i_2}(||X_{i_1i_2}||-||\bar{X}_{i_1i_2}||)^2, \\
&=\sum_{(i_1i_2)}S_{i_1i_2}( ||\bar{X}_{i_1i_2}||^2-2||X_{i_1i_2}||\,||\bar{X}_{i_1i_2}||)+c, \nonumber
\end{align}
where $c$ is a constant and ${S}_{i_1i_2}$ measures the weight of $(||X_{i_1i_2}||-||\bar{X}_{i_1i_2}||)^2$ if we have priors. We denote $\mathbf{S} \triangleq \{ S_{i_1i_2} \}\in \mathbb{R}^{m\times m}$.

The gradient of $F_{XY}(\mathbf{P})$ w.r.t. $\mathbf{P}$ can be computed using the chain rule,
\begin{equation}\label{gra_F}
\nabla F_{XY}(\mathbf{P}) = 2(\mathbf{L}_X+\mathbf{L}^*_X)(\mathbf{P}Y)Y^T,
\end{equation}
where $\bf{L}_X=\text{diag}(\bf{SI}_m)-S$ is the Laplacian of $\mathcal{G}_X$, and $\mathbf{L}^*_X=\text{diag}(\bf{S^*I}_m)-S^*$ with $S_{i_1i_2}^*\triangleq S_{i_1i_2}||X_{i_1i_2}||\,||\bar{X}_{i_1i_2}||^{-1}$. To avoid numerical instabilities as in~\cite{[2007-Candès]}, a small $\epsilon > 0$ is added to $||\bar{X}_{i_1i_2}||^{-1}$, {\it i.e.}, $(||\bar{X}_{i_1i_2}||+\epsilon)^{-1}$. Naturally, we can reconstruct $F_{XY}(\mathbf{P})$ by adding a unary potential such as $\sum_{ij} f_v(X_i,Y_j)\mathbf{P}_{ij} + \lambda F_{XY}(\mathbf{P})$.

Due to the non-convexity of $F_{XY}(\mathbf{P})$, its minimizer $\mathbf{P}^*\in \hat{\mathcal{P}}$, which is regarded as an optimal transformation map from $\mathcal{G}_X$ to $\mathcal{G}_{\bar{X}}$, often reaches a local minimum and is not binary; see Fig.\ref{fig:longfig1} (b) for illustration. Consequently, the transformed node $\bar{X}_{i}$ is usually not exactly equal to a $Y_{j'}\in Y$, and there is often a shift between $\bar{X}_i$ and its correct match $Y_{\sigma_i}$. Fig.\ref{fig:longfig1} (a) displays this shift phenomena, where each $\bar{X}_i$ shifts from the correct match $Y_{\sigma_i}$ to some degree.

\subsection{Node shifting}
Benefiting from the property of $F_{XY}(\mathbf{P})$ that preserves the edge alignment between $\mathcal{G}_X$ and $\mathcal{G}_{\bar{X}}$, the shifting vectors of adjacent nodes have similar directions and norms, as shown in Fig.\ref{fig:longfig1}~(a). Consequently, in order to reduce the node shifting from $\bar{X}_i$ to its correct match $Y_{\sigma_i}$, denoted by
$$
\overrightarrow{\bar{X}_{i}Y_{\sigma_i}}=Y_{\sigma_i}-\bar{X}_{i},
$$
we minimize the sum of the differences between adjacent shifting vectors, {\it i.e.},
\begin{align}
G_{\bar{X}Y}(\mathbf{P})&=\sum_{(i_1i_2)}\bar{S}_{i_1,i_2}||(\bar{\bar{X}}_{i_1}-\bar{X}_{i_1})-(\bar{\bar{X}}_{i_2}-\bar{X}_{i_2})||_2^2\nonumber\\
&=\text{Tr}\left((\mathbf{P}Y-\bar{X})^T\mathbf{L}_{\bar{X}}(\mathbf{P}Y-\bar{X})\right),
\end{align}
where $\mathbf{L}_{\bar{X}} = \text{diag}(\bar{\mathbf{S}}\mathbf{I}_m)-\bar{\mathbf{S}}$. We denote $\bar{\bar{X}}=\mathbf{P}Y$ as the transformed nodes of $\bar{X}$. In our method, the weight matrix $\bar{\mathbf{S}}$ is set to be positive and symmetric, therefore, $\mathbf{L}_{\bar{X}}$ is positive definite and $G_{\bar{X}Y}(\mathbf{P})$ is convex.

{\bf Sparse regularization} 
Because $G_{\bar{X}Y}(\mathbf{P})$ is convex, its minimizer is often an inner point rather than an extreme point of the feasible field $\hat{\mathcal{P}}$. In order to approach a binary solution, we first add a sparse regularization term, {\it i.e.}, the $l_1$ norm of $\mathbf{P}$ to $G_{\bar{X}Y}(\mathbf{P})$.
We denote $D_{ij} \triangleq d(\bar{X}_i,Y_j)$ as the distance between $\bar{X}_i$ and $Y_j$.
Benefiting from the solution of $F_{XY}(\mathbf{P})$, the norms of shifting vectors $\overrightarrow{\bar{X}_{i}Y_{\sigma_i}}$ are relatively small, and elements $D_{i,\sigma_i}$ are much smaller than $D_{i,j\neq \sigma_i}$, as shown in Fig.\ref{fig:1} (f).
Thus, we also add a unary term $\mathbf{D}_{\bar{X}Y}=\left\{ D_{ij}\right\} \in \mathbb{R}^{m\times n}$ to improve the sparsity of the minimizer.

Finally, $G_{\bar{X}Y}(\mathbf{P})$ can be summarized as
\begin{align}
G_{\bar{X}Y}(\mathbf{P})=\langle\mathbf{P},\mathbf{D}_{\bar{X}Y}\rangle + \lambda_1 ||\mathbf{P}||_1^1\nonumber+\lambda_2\text{Tr}\left((\mathbf{P}Y-\bar{X})^T\mathbf{L}_{\bar{X}}(\mathbf{P}Y-\bar{X})\right),
\end{align}
where $\langle\mathbf{P},\mathbf{D}_{\bar{X}Y}\rangle=\sum_{ij}\mathbf{P}_{ij}D_{ij}$. The gradient of $G_{\bar{X}Y}(\mathbf{P})$ is then
\begin{equation}\label{gra_G}
\nabla G_{\bar{X}Y}(\mathbf{P}) =  \mathbf{D}_{\bar{X}Y} + \lambda_1 + 2\lambda_2\mathbf{L}_{\bar{X}}(\mathbf{P}Y-\bar{X})Y^T.
\end{equation}
With this sparse regularization, the function $G_{\bar{X}Y}(\mathbf{P})$ is always solved with a (nearly) binary solution, which significantly improves the matching accuracy. See Fig.~\ref{fig:longfig1}~(c) and (d) for examples.

\subsection{Outlier removal  via domain adaptation}\label{section33}

\begin{figure}[htb!]
	\begin{center}
		\subfigure
		{\includegraphics[width=0.16\linewidth]{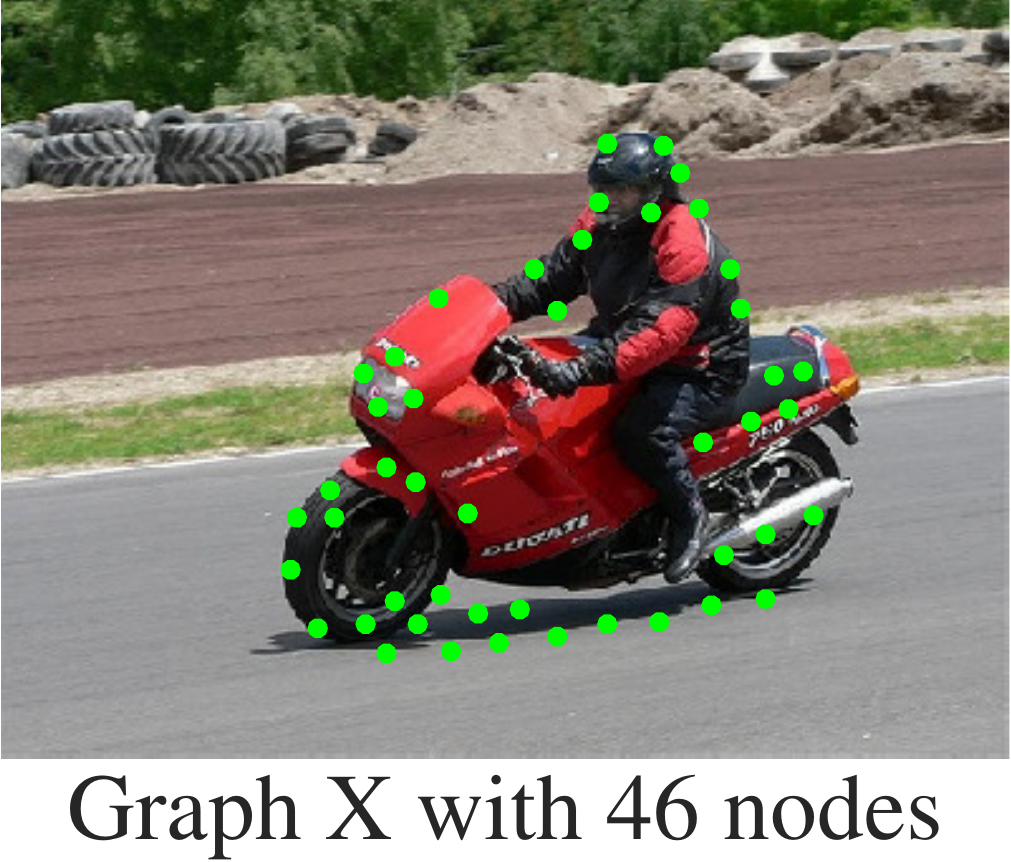}}
		\subfigure
		{\includegraphics[width=0.16\linewidth]{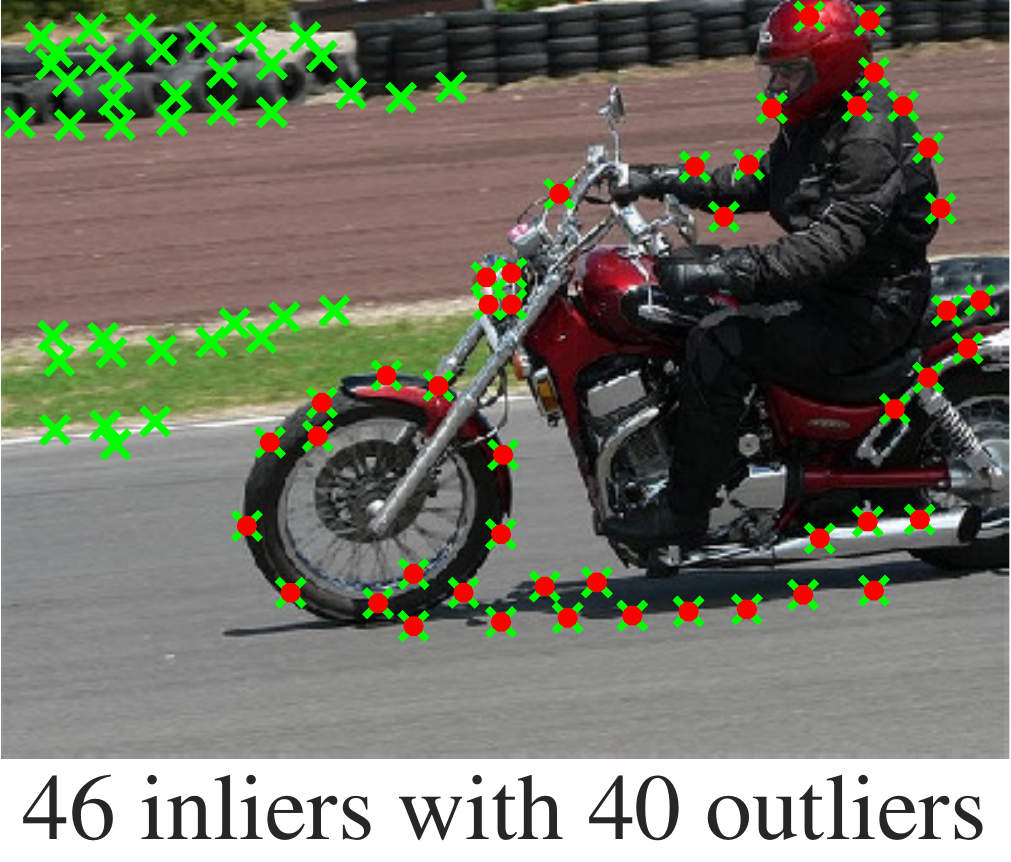}}
		\subfigure
		{\includegraphics[width=0.16\linewidth]{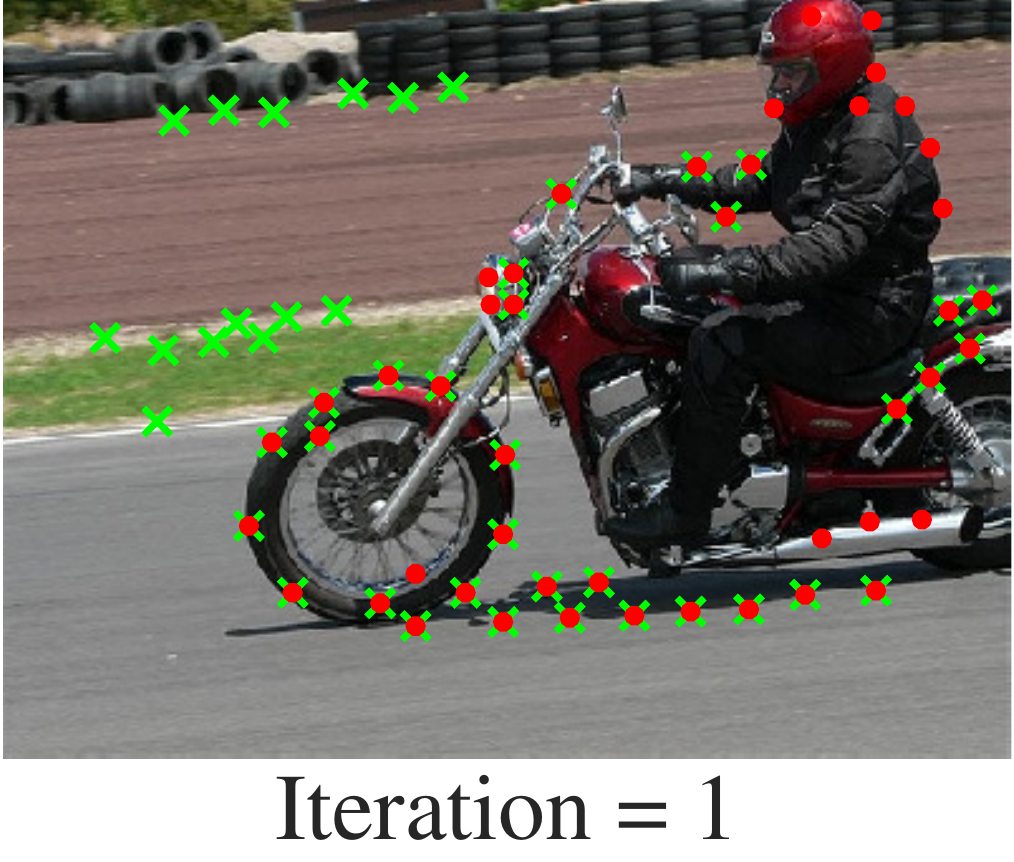}}
		\subfigure
		{\includegraphics[width=0.16\linewidth]{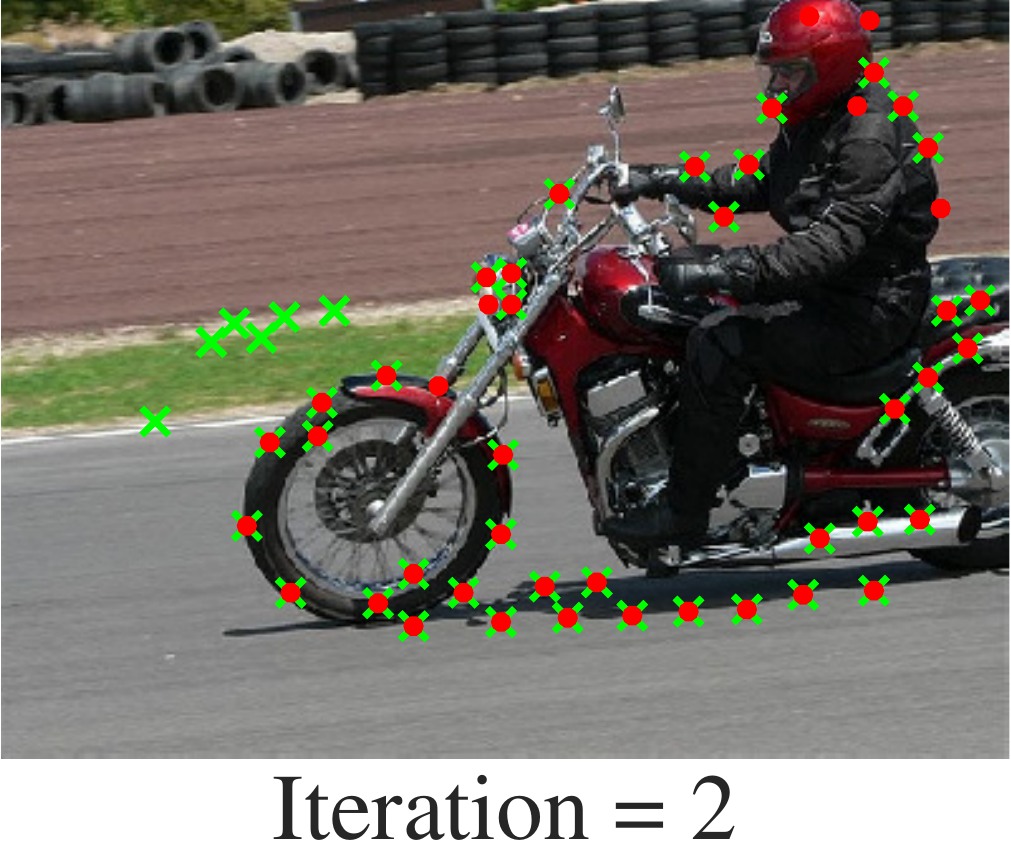}}
		\subfigure
		{\includegraphics[width=0.16\linewidth]{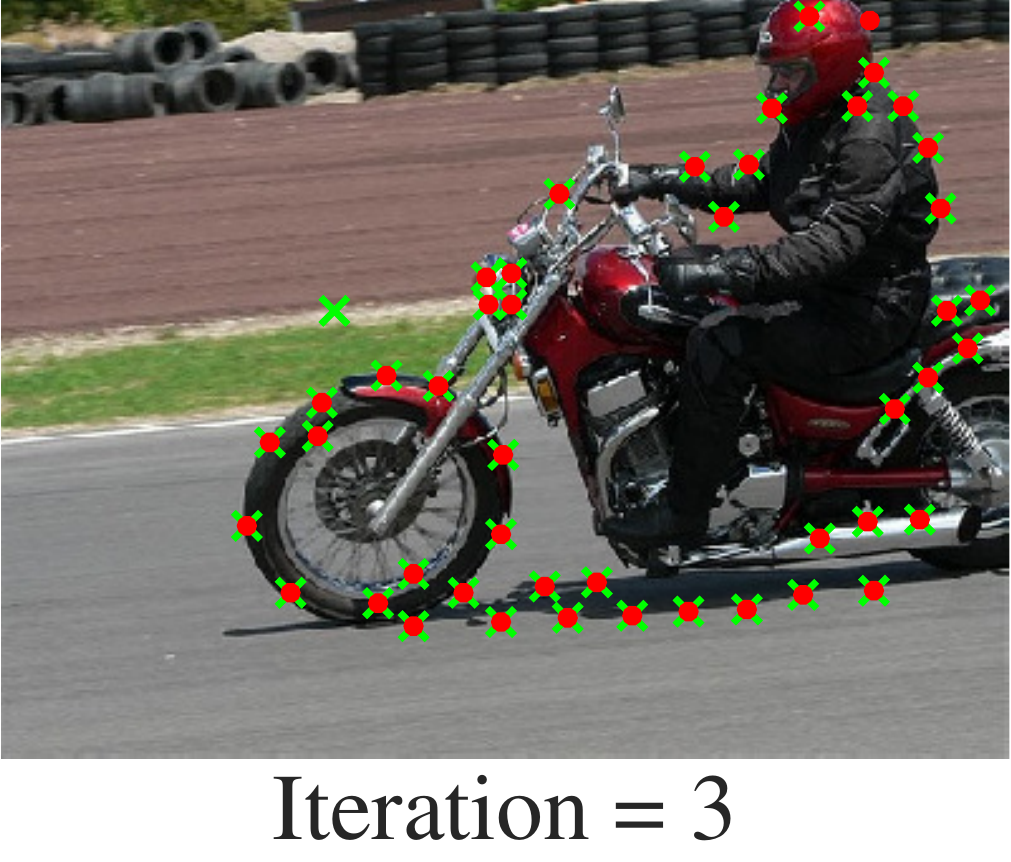}}
		\subfigure
		{\includegraphics[width=0.16\linewidth]{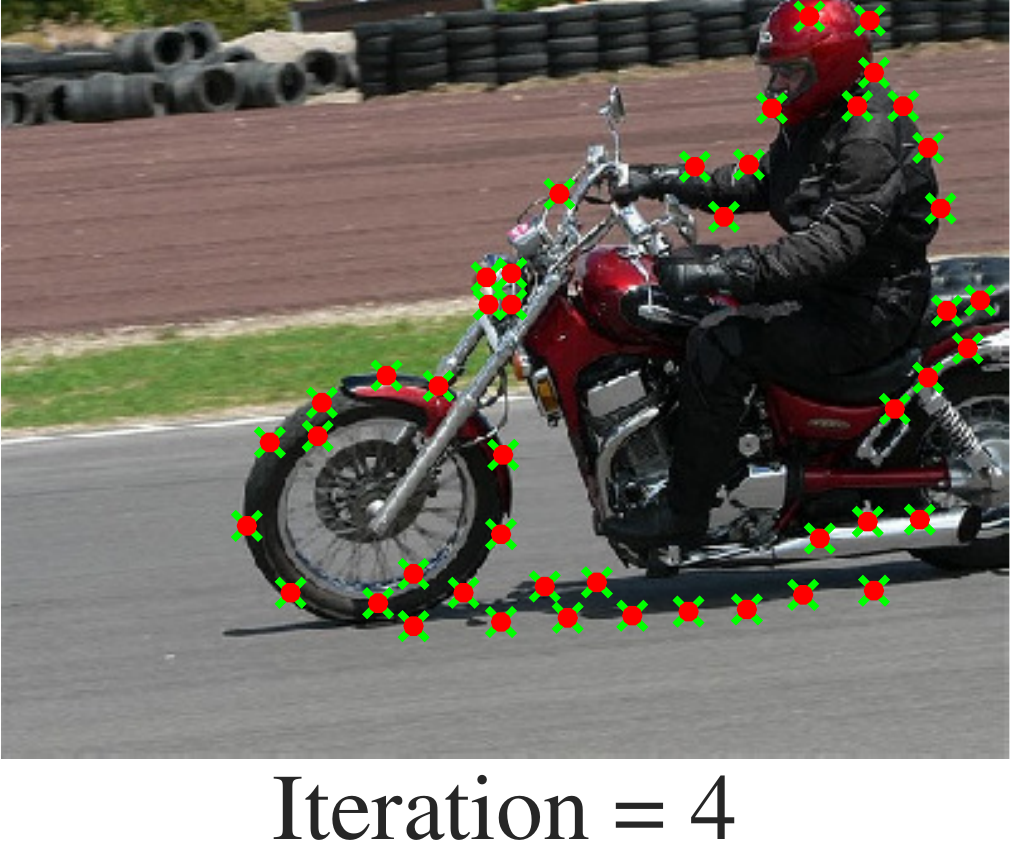}}
	\end{center}
	\caption{Outlier removal with a transformation map $\mathbf{P}^*$ obtained by alternately minimizing $F_{XY}(\mathbf{P})$ and $G_{XY}(\mathbf{P})$. In each iteration, the red dots are inliers and the green plus signs are the nodes remaining after removal.}
	\label{fig:removal}
\end{figure}

Matching graphs $\mathcal{G}_X$ and $\mathcal{G}_Y$ of different sizes with $m<n$ is more complicate. In this situation, the outliers occurring in graph $\mathcal{G}_Y$ usually affect the matching results. Thanks to the transformation map $\mathbf{P}^*$ achieved by minimizing $F_{XY}(\mathbf{P})$, the structure of $\mathcal{G}_{\bar{X}}$ is similar to that of $\mathcal{G}_{X}$. In some sense, the operation $\bar{X}=\mathbf{P}^*Y$ can be seen as a domain adaptation~\cite{[2017-Courty-pami]} from the source domain $X$ to the target domain $Y$. We propose a method to remove outliers adaptively by using the transformation map alternately minimized from $F_{XY}(\mathbf{P})$ and $G_{XY}(\mathbf{P})$, where $G_{XY}(\mathbf{P})$ is defined by replacing $\bar{X}$ with $X$ in the pairwise potential of $G_{\bar{X}Y}(\mathbf{P})$:
\begin{align}
G_{XY}(\mathbf{P})&=\sum_{(i_1,i_2)}S_{i_1i_2}||(\bar{X}_{i_1}-X_{i_1})-(\bar{X}_{i_2}-X_{i_2})||_2^2\\
&=\sum_{(i_1,i_2)}S_{i_1i_2}||(X_{i_2}-X_{i_1})-(\bar{X}_{i_2}-\bar{X}_{i_1})||_2^2,
\end{align}
which depicts the edge-vector differences between the original graph $\mathcal{G}_X$ and the transformed graph $\mathcal{G}_{\bar{X}}$.
The orientation of edge has also been used in many graph matching methods ~\cite{[2017-Huu-cvpr],[2013-Torresani-pami],[2016-Zhou-pami]} to construct the non-diagonal element of the affinity matrix as:
\begin{equation}
\small{
	\mathbf{W}_{i_1j_1,i_2j_2}=\text{exp}\left(-\frac{1}{2}(||X_{i_1i_2}||-||Y_{j_1j_2}||)^2-\frac{1}{2}(\theta_{i_1i_2}-\theta_{j_1j_2})^2\right),}
\label{equationedge}
\end{equation}
where $\theta_{i_1i_2}$ is the angle between edge $X_{i_1i_2}$ and the horizontal line.

After minimizing $F_{XY}(\mathbf{P})$ or $G_{XY}(\mathbf{P})$, we obtain the transformed nodes $\bar{X}=\mathbf{P}^*Y$. Consequently, $\mathcal{G}_{\bar{X}}$ has a structure similar to that of the original graph $\mathcal{G}_X$ and lies in the same coordinate system of $\mathcal{G}_Y$ with relatively small shifts. Then, we can remove outliers adaptively using a ratio test technique. Given two point sets $\bar{X}$ and $Y$, we compute the Euclidean distance $d_{ij}$ between all the pairs $(\bar{X}_i,Y_{j})$.  For each node $\bar{X}_i$, we find the closest node $Y_{j^*}$ and remove all the nodes $Y_j$ when $d_{ij}>k\cdot d_{ij^*}$ for a given $k>0$. If the number  of remaining nodes $l$ is smaller than $m$, $m-l$ nodes are selected from the removed nodes that are closer to $\bar{X}$ and are added. The experimental results show that after several iterations of alternately minimizing $F_{XY}(\mathbf{P})$ and $G_{XY}(\mathbf{P})$ most outliers are removed (see Fig.\ref{fig:removal}). 

Our ATGM algorithm with outlier-removal is summarized in {\bf{Algorithm}~\ref{algorithm}}.
\begin{algorithm}
	\caption{~~$\mathbf{P}^*\leftarrow {\bf{ATGM}}(X,Y,k_0)$}
	\begin{algorithmic}
		\STATE {\bf{Input~~~:}} $X,~Y,~k_0$ and $\mathbf{S}, \bar{\mathbf{S}}$ if available.
		\STATE {\bf{Output:}} $\mathbf{P}^*$
		\WHILE{$k\le k_0$}
		\STATE $ \mathbf{P}^*\leftarrow \text{argmin} ~ G_{XY}$ via Eq.\eqref{eq:fw1} and Eq.\eqref{eq:fw2};
		\STATE $Y~~\leftarrow$ removing outliers of $Y$ with $\mathbf{P}^*$;
		\STATE $\mathbf{P}^*\leftarrow \text{argmin} ~ F_{XY}$ via Eq.\eqref{eq:fw1} and Eq.\eqref{eq:fw2};
		\STATE $Y~~\leftarrow$ removing outliers of $Y$ with $\mathbf{P}^*$;
		\STATE $k ~~~\leftarrow k+1$;
		\ENDWHILE
		\STATE ~~~~$\mathbf{P}^*\leftarrow \text{argmin} ~ F_{XY}$ via Eq.\eqref{eq:fw1} and Eq.\eqref{eq:fw2};
		\STATE ~~~~$\bar{X}~\leftarrow\mathbf{P}^*Y$;
		\STATE ~~~~$ \mathbf{P}^*\leftarrow \text{argmin} ~ G_{\bar{X}Y}$ with $\mathbf{P}^* $ as initialization.
		\STATE ~~~~$\mathbf{P}^*\leftarrow $ post-discretization of $ \mathbf{P}^* $ by the Hungarian method.
	\end{algorithmic}
	\label{algorithm}
\end{algorithm}

\section{Numerical implementation and analysis}
As presented above, we construct two objective functions, namely, a non-convex $F_{XY}(\mathbf{P})$ and a convex $G_{\bar{X}Y}(\mathbf{P})$. 
Previous methods, {\itshape e.g.},~\cite{[2009-Zaslavskiy-pami],[2016-Zhou-pami],[2014-Liu-pami]}, relaxed their objective functions in both convex and concave forms as $\mathbf{J}_v$ and $\mathbf{J}_c$, respectively, and solved a series of combined functions $\mathbf{J}_{\lambda}=\lambda \mathbf{J}_c + (1-\lambda)\mathbf{J}_v$ controlled by a parameter $\lambda$ increasing from $0$ to $1$. In contrast, we solve our objective functions $F_{XY}(\mathbf{P})$ and $G_{\bar{X}Y}(\mathbf{P})$ separately by the Frank-Wolfe (FW) method~\cite{[2009-Zaslavskiy-pami],[2016-Zhou-pami]}, which is simple but efficient. 

Given that $g$ is a convex and differentiable function, and given that $\hat{\mathcal{P}}$ is  a convex set, the FW method iterates the following steps until it converges:
\begin{eqnarray}
&& \tilde{\mathbf{P}}^{(k+1)}\in \mathop{\text{argmin}}\limits_{\mathbf{P}\in \hat{\mathcal{P}}}\langle\nabla g(\mathbf{P}^{(k)}),\mathbf{P}\rangle\label{equation18},
\label{eq:fw1}
\\
&& \mathbf{P}^{(k+1)}=\mathbf{P}^{(k)} + \alpha^{(k)}(\tilde{\mathbf{P}}^{(k+1)}-\mathbf{P}^{(k)}),
\label{eq:fw2}
\end{eqnarray}
where $\alpha^{(k)}$ is the step size of the iteration $k$ obtained by a line search procedure~\cite{[Goldstein-1965]}, and $\nabla g$ is computed using the Eq.\eqref{gra_F} and Eq.\eqref{gra_G}.

In Eq.\eqref{eq:fw1}, the minimizer $\tilde{\mathbf{P}}^{(k+1)} \in \hat{\mathcal{P}}$ is theoretically an extreme point of $\hat{\mathcal{P}}$ (so is binary). This means that $\tilde{\mathbf{P}}^{(k+1)} \in \mathcal{P}$. Therefore, Eq.\eqref{eq:fw1} is a linear assignment problem (LAP) that can be efficiently solved by approaches such as the Hungarian~\cite{[2010-Kuhn]}, LAPJV~\cite{[1987-Jonker]} algorithm. Moreover, since $\tilde{\mathbf{P}}^{(k+1)}$ is binary in each iteration, the final solution $\mathbf{P}^*$ is (nearly) binary after minimizing $G_{\bar{X}Y}(\mathbf{P})$.

{\bf Convergence} The FW method ensures an at least sublinear convergence rate~\cite{[2015-Simon-nips]}, which may result in large iterations for solving the non-convex function $F_{XY}(\mathbf{P})$. However, minimizing $F_{XY}(\mathbf{P})$ within 200 iterations is sufficient because its solution will be applied as the initialization for minimizing $G_{\bar{X}Y}(\mathbf{P})$, which is strong convex and stronger convergence can be achieved. In our experiments, $G_{\bar{X}Y}(\mathbf{P})$ always converges at a $10^{-7}$ tolerance within $k\le 100$ iterations. Compared to the path-following method that solves the two relaxed objective functions combined together in~\cite{[2009-Zaslavskiy-pami],[2016-Zhou-pami],[2014-Liu-pami]}, our optimization strategy is faster with higher matching accuracy.

{\bf Local optimal vs. global optimal} The FW method can guarantee obtaining only a local optimum of the non-convex objective $F_{XY}(\mathbf{P})$. However, as discussed above, the local optimum for $F_{XY}(\mathbf{P})$ is applied as an initialization for solving the convex objective $G_{\bar{X}Y}(\mathbf{P})$, which allows us to reach a global optimum.

{\bf Computational complexity}\label{complexity} For our method, the space complexity is $\mathbf{O}(mn)$, which is considerably smaller than the size $\mathbf{O}(m^2n^2)$ of most of other methods with complete graphs. The time complexity is $\mathbf{O}(Tn^3)$, where $T$ is the number of iterations in the FW method. This complexity can be calculated as $\mathbf{O}\left(T(\tau_{f} +\tau_{l}) + \tau_s \right)$, where $\tau_s=\mathbf{O}(m^2)$ is the cost of the edge attribute matrices of $\mathcal{G}_X$. In each iteration of the FW method, $\tau_{f}=\mathbf{O}(m^2n)$ is the cost to compute the gradient, function value and step size at $\mathbf{P}^{(k)}$, and $\tau_{l}=\mathbf{O}(n^3)$ is the cost to minimize Eq.\eqref{eq:fw1} using the Hungarian algorithms.

\section{Experimental analysis}\label{section_results}
In this section, we evaluate our method {\bf{ATGM}} on both synthetic data and real-world datasets. We compare our method with state-of-the-art methods including GA~\cite{[1996-Gold]}, PM~\cite{[2008-Zass-cvpr]}, SM~\cite{[2005-Leordeanu]}, SMAC~\cite{[2006-Cour-nips]}, IPFP~\cite{[2009-Leordeanu-nips]}, RRWM~\cite{[2010-Cho-eccv]}, FGM~\cite{[2016-Zhou-pami]} and MPM~\cite{[2014-Cho-cvpr]}. As suggested in ~\cite{[2009-Leordeanu-nips]}, we use the solution of SM as the initialization for IPFP. Also, for FGM, we use the deformable graph matching method called FGM-D.

In all the experiments, to be able to apply unified parameters $\lambda=1,\lambda_1=10^3$, and $\lambda_2=1$, we normalize the node coordinates to $[0,1]$ for our method. For the non-convex objective functions $F_{XY}$, we compute its unary term by using Shape Context~\cite{[2002-Belongie-pami]}. For comparison, the average accuracy for each algorithm is reported. Our objective functions are different from those used by the compared methods, and thus, it does not make sense to  compare the objective scores or objective ratios.

\subsection{Results on synthetic data}

We perform a comparative evaluation of ATGM on  synthesized random point sets following  ~\cite{[2016-Zhou-pami],[2017-Jiang-cvpr],[2010-Cho-eccv]}. The synthetic points of $\mathcal{G}_X$ and $\mathcal{G}_Y$ are constructed as follows: for the graph $\mathcal{G}_X$, $n_{in}$  inlier points are randomly generated on $\mathbb{R}^2$ with the Gaussian distribution $\mathcal{N}(0,1)$. 
The graph $\mathcal{G}_Y$ with noise is generated by adding Gaussian noise $\mathcal{N}(0,\sigma^2)$ to each $X_i\in X$ to evaluate the robustness of the method to deformation noise. 
Graph $\mathcal{G}_Y$ with outliers is generated by adding $n_{out}$ additional points on $\mathbb{R}^2$ with a Gaussian distribution $\mathcal{N}(0,1)$ to evaluate the robustness to outliers.

For the  compared methods, as in~\cite{[2016-Zhou-pami]}, we set the edge affinity matrix $\mathbf{W}_{i_1j_1,i_2j_2}=\text{exp}(-\frac{(||X_{i_1i_2}||-||Y_{j_1j_2}||)^2}{0.15})$ . We set $\mathbf{S}\in \mathbb{R}^{m\times m}$ as $S_{i_1i_2}={||X_{i_1i_2}||}^{-1}$  for $F_{XY}(\mathbf{P})$ an $G_{XY}(\mathbf{P})$ with fully connected $\mathcal{G}_X$. For $G_{\bar{X}Y}(\mathbf{P})$, our method performs a Delaunay triangulation on $X$ to get its edge set $\bar{\mathcal{E}}_{X}$, and then, $\bar{\mathcal{E}}_{X}$ is divided into two parts using k-means by considering the edge length (edges with longer lengths are abandoned).

{\bf Memory efficiency}
As analyzed in Sec.\ref{complexity}, the space complexity of our method is lower than that of compared methods.
In this experiment, we try to verify that ATGM can match graphs with low memory consumption while achieving better accuracy. 

Since the compared methods can achieve better accuracy with complete graphs, for fairness, we first applied all methods to complete graphs with a relative small size $n_{in}=20$.  We then enlarged the size to $n_{in}=100$ to test the advantages of ATGM in terms of memory efficiency. Due to the high space complexities of the other methods, we had to apply them to graphs with Delaunay triangulation. However, our method is able to use complete graphs due to its lower space complexity $O(n^2)$.

\begin{figure}[htb!]
\centering
		{\includegraphics[width=0.8\linewidth]{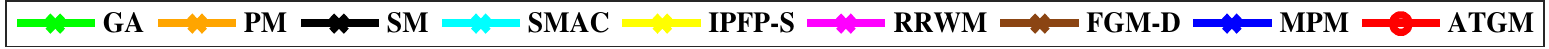}}\\
		\subfigure[]
		{\includegraphics[width=0.24\linewidth]{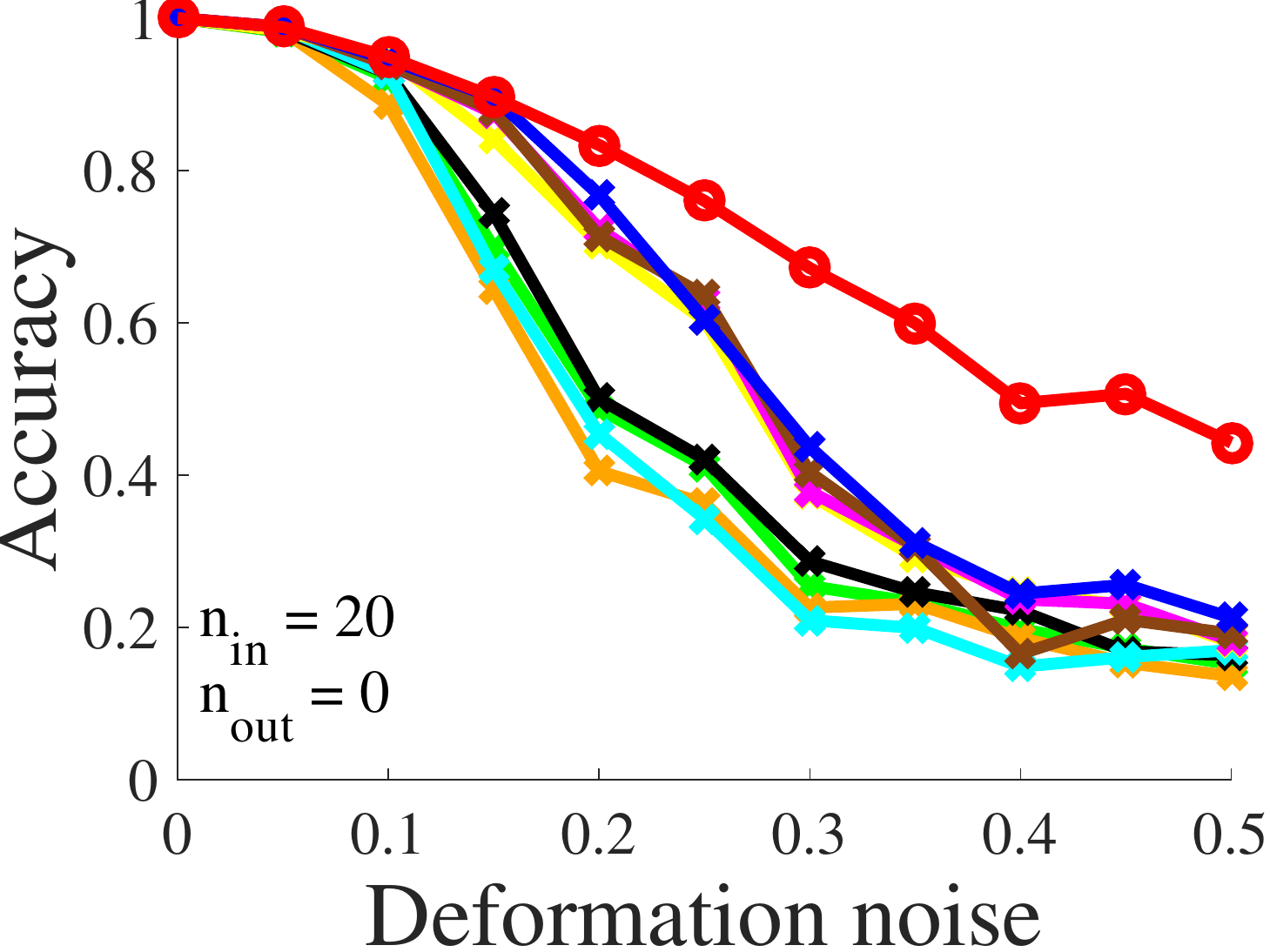}}
		\subfigure[]
		{\includegraphics[width=0.24\linewidth]{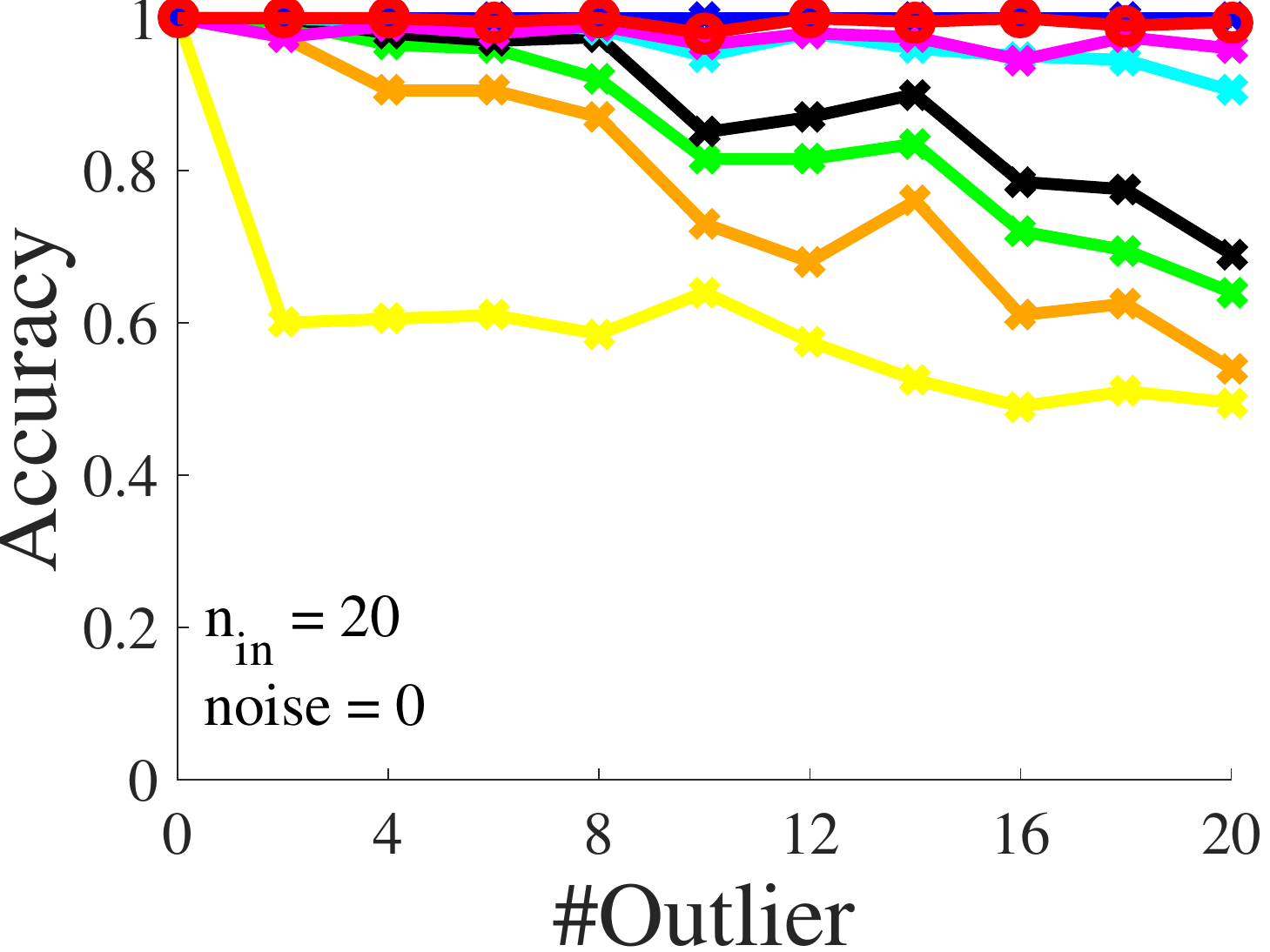}}
		\subfigure[]
		{\includegraphics[width=0.24\linewidth]{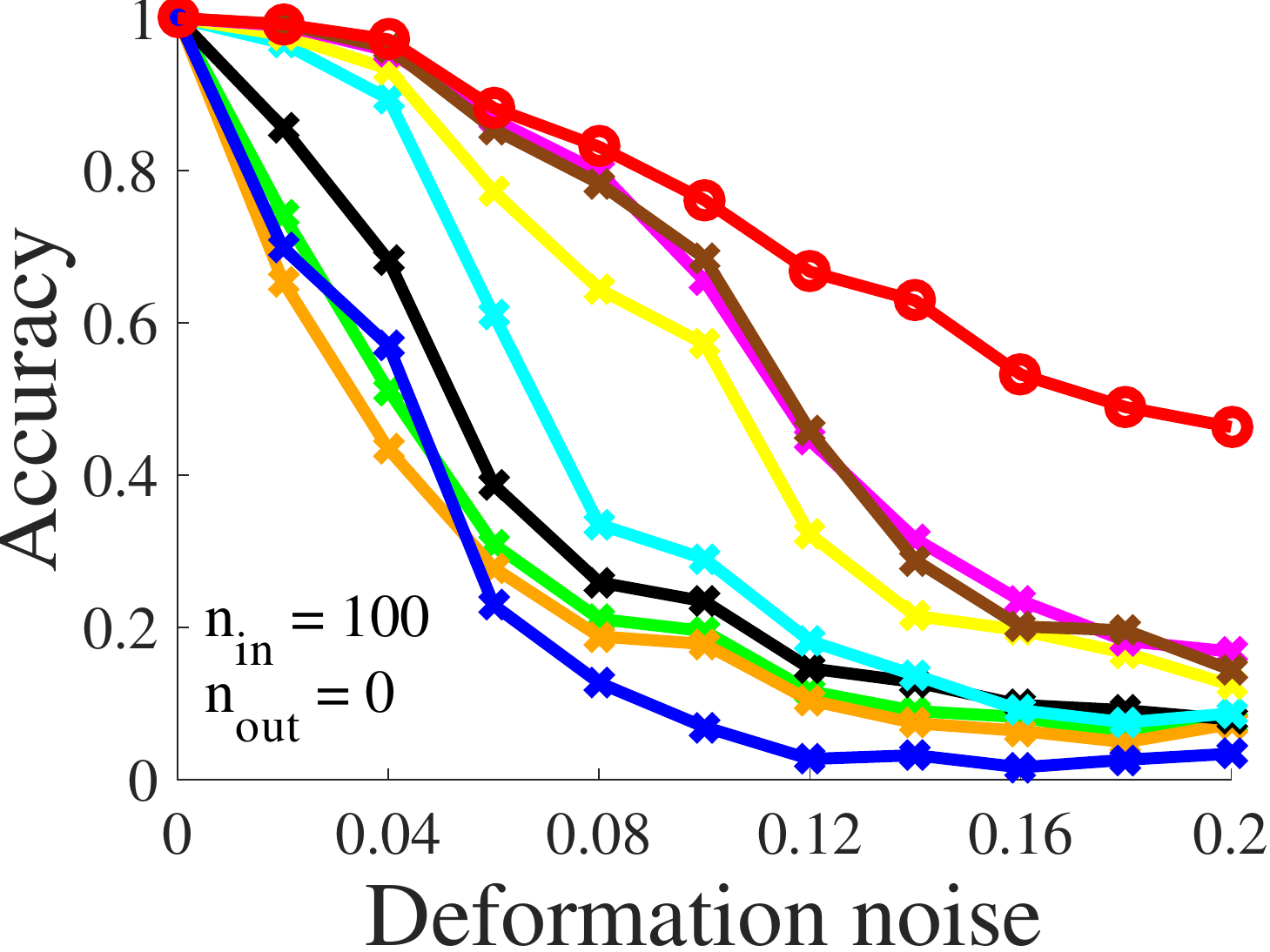}}
		\subfigure[]
		{\includegraphics[width=0.24\linewidth]{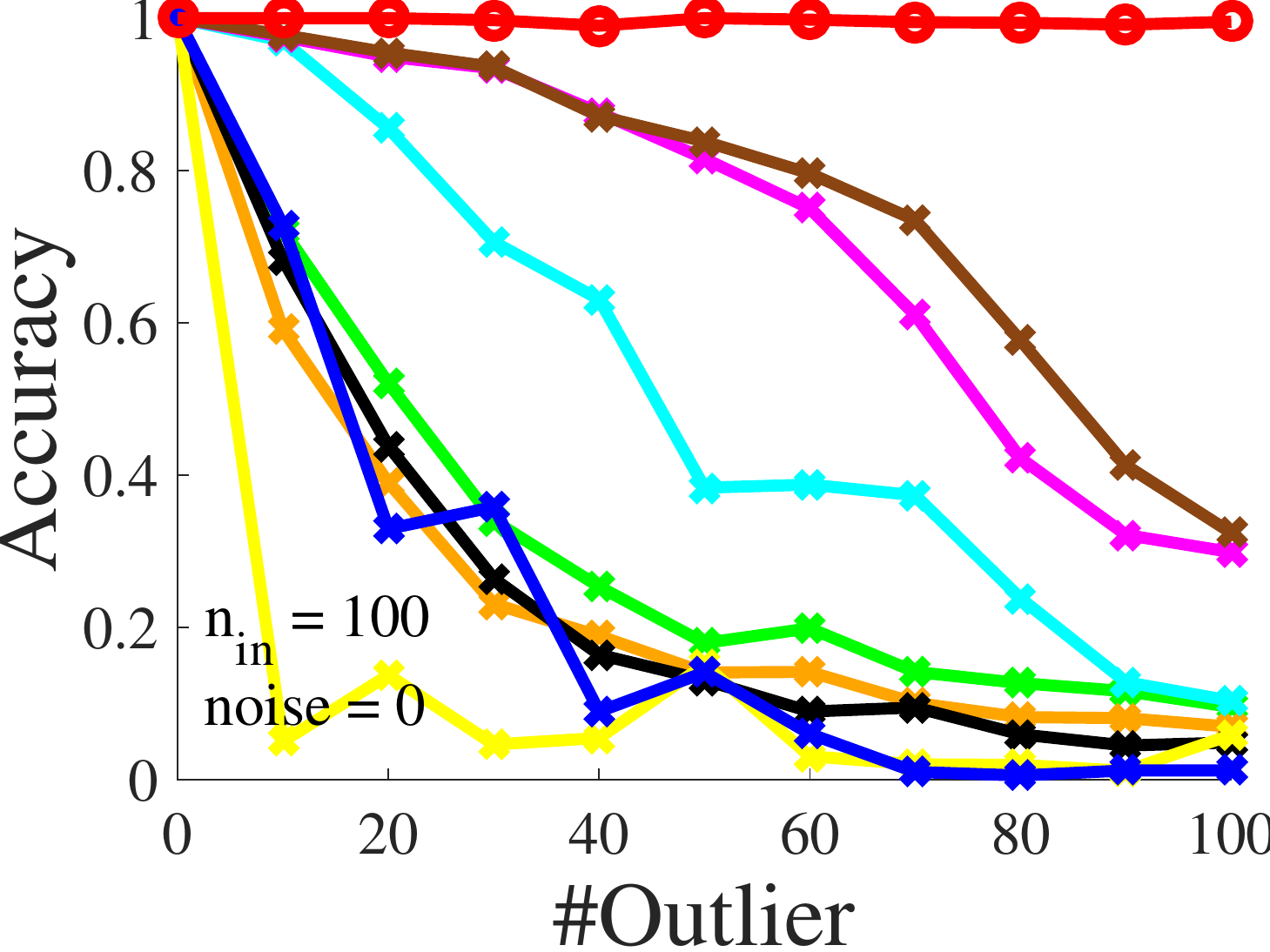}}
	\caption{Comparisons of the robustness to noise and outliers. For complete graphs, the accuracy with respect to the noise and number of outliers are in (a) and (b), respectively. 
		The results for graphs connected by Delaunay triangulation are shown in (c) and (d).}
	\label{fig:syn_menory}
\end{figure}

As shown in Fig.\ref{fig:syn_menory} (a) and (b), under the complete graph setting, our method achieves the highest average accuracy in the case with deformation noise and achieves competitive results in the case with outliers. 
For graphs of large size, our method outperforms all the other methods (shown in Fig.\ref{fig:syn_menory} (c) and (d)). In contrast, using complete graphs with a large number of nodes with other methods is infeasible in practice. Except for PM~\cite{[2008-Zass-cvpr]}, all of the compared methods have to use  $n_{in}^2(n_{in}+n_{out})^2$  units of memory , which will be extremely large for $n_{in}=100, n_{out}\geq 100$.  This requirement affects their application to graph matching in practice.
	
{\bf Running time} To compare the time consumption of all methods, we tested them in both equal-size and unequal-size cases, namely, (1) $n_{in}=10,20,...,100$, $n_{out}=0$, $\sigma = 0.2$ and (2) $n_{in} = 100$, $n_{out} = 10,20,...,100$, $\sigma = 0.05$. Considering the effect of the number of edges on time consumption, in equal-size cases, we applied all methods to both complete and Delaunay triangulation-connected graphs. In unequal-size cases, we applied our method to complete graphs and the others to Delaunay triangulation-connected graphs so that ATGM took more edges than the others.

\begin{figure}[htb!]
	\centering
	{\includegraphics[width=0.8\linewidth]{./Images/fig_henglan.pdf}}\\
	{\includegraphics[width=0.3\linewidth]{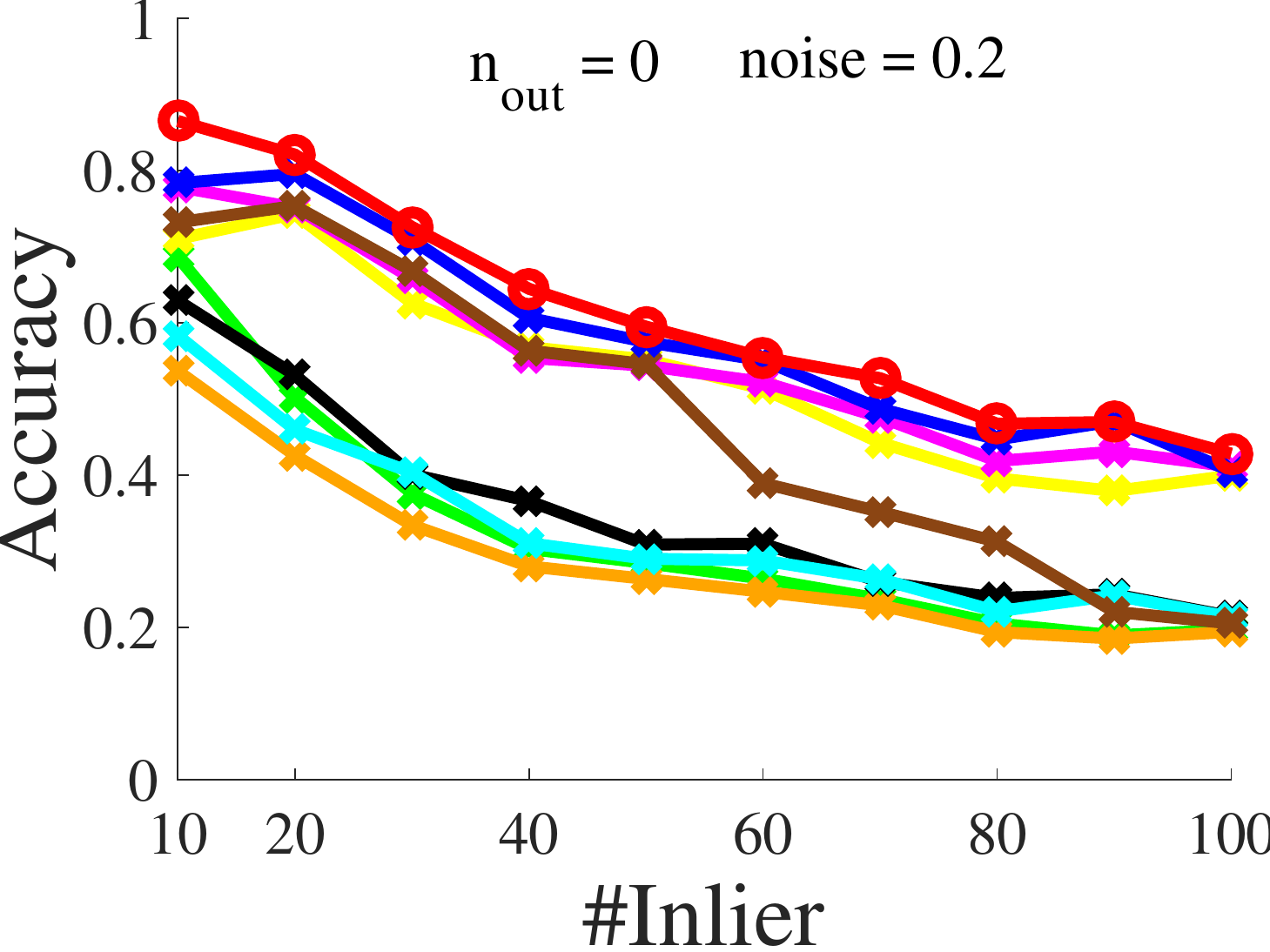}}
	{\includegraphics[width=0.3\linewidth]{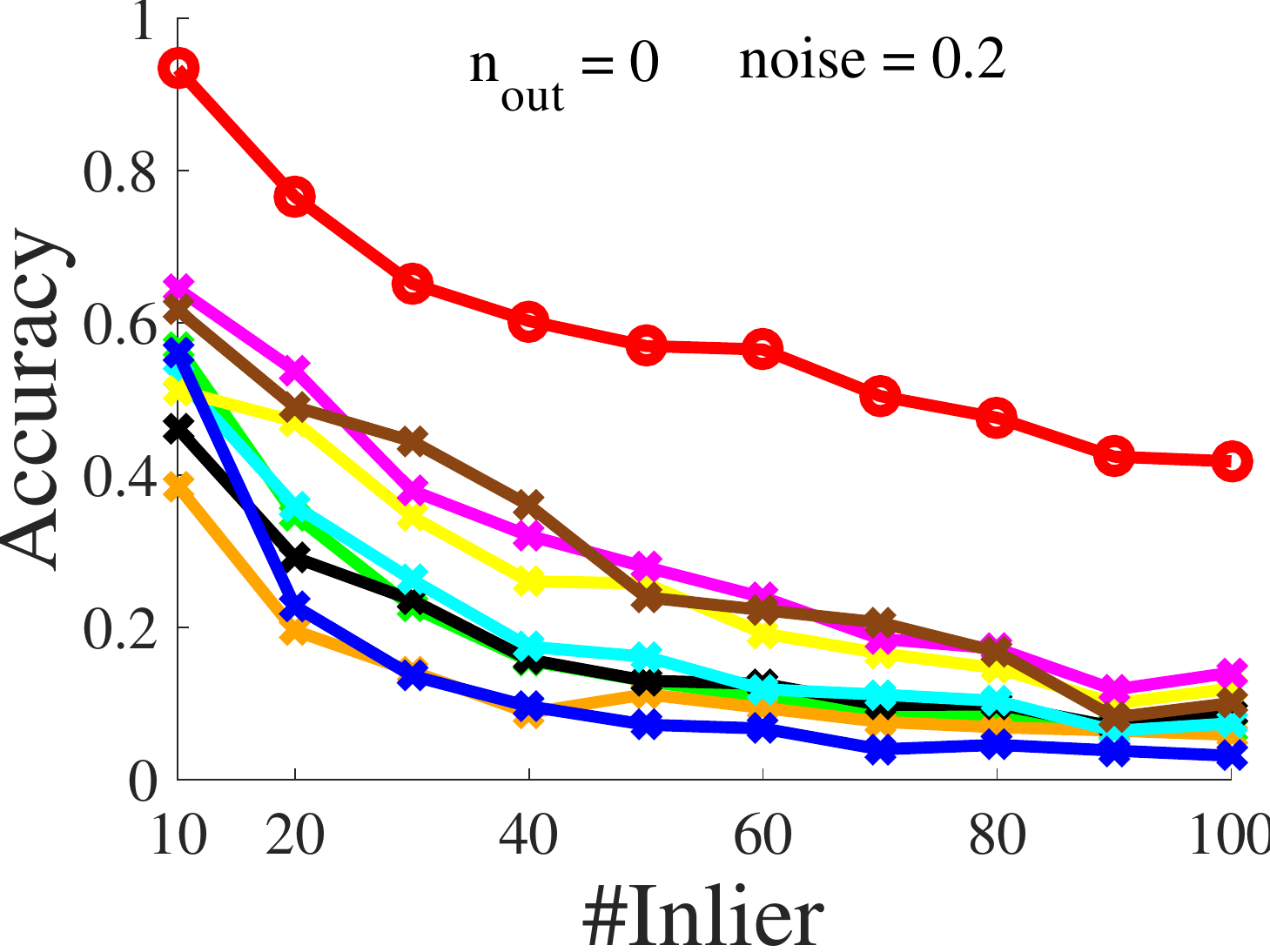}}
	{\includegraphics[width=0.3\linewidth]{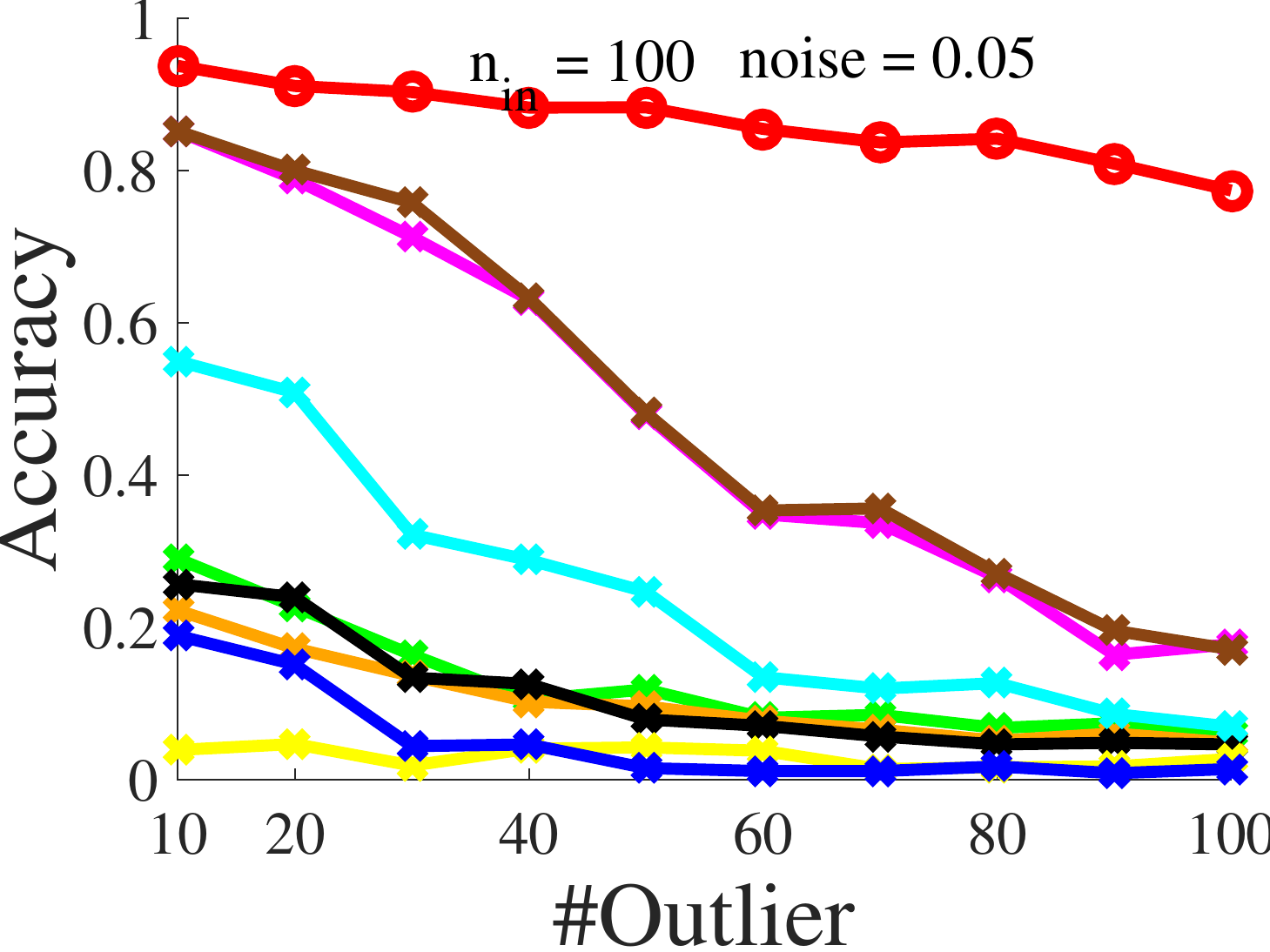}}		
	\subfigure[]
	{\includegraphics[width=0.3\linewidth]{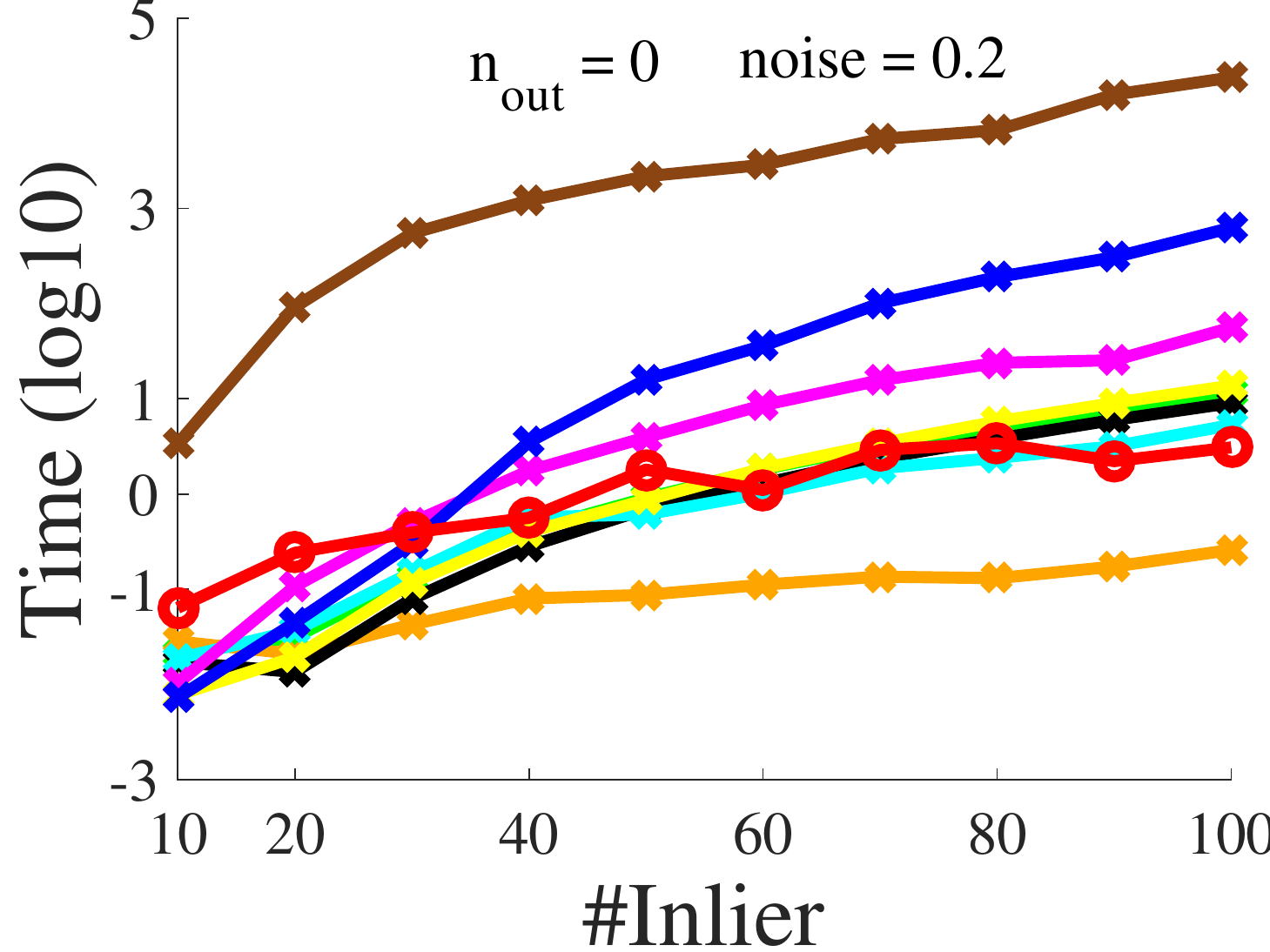}}	
	\subfigure[]
	{\includegraphics[width=0.3\linewidth]{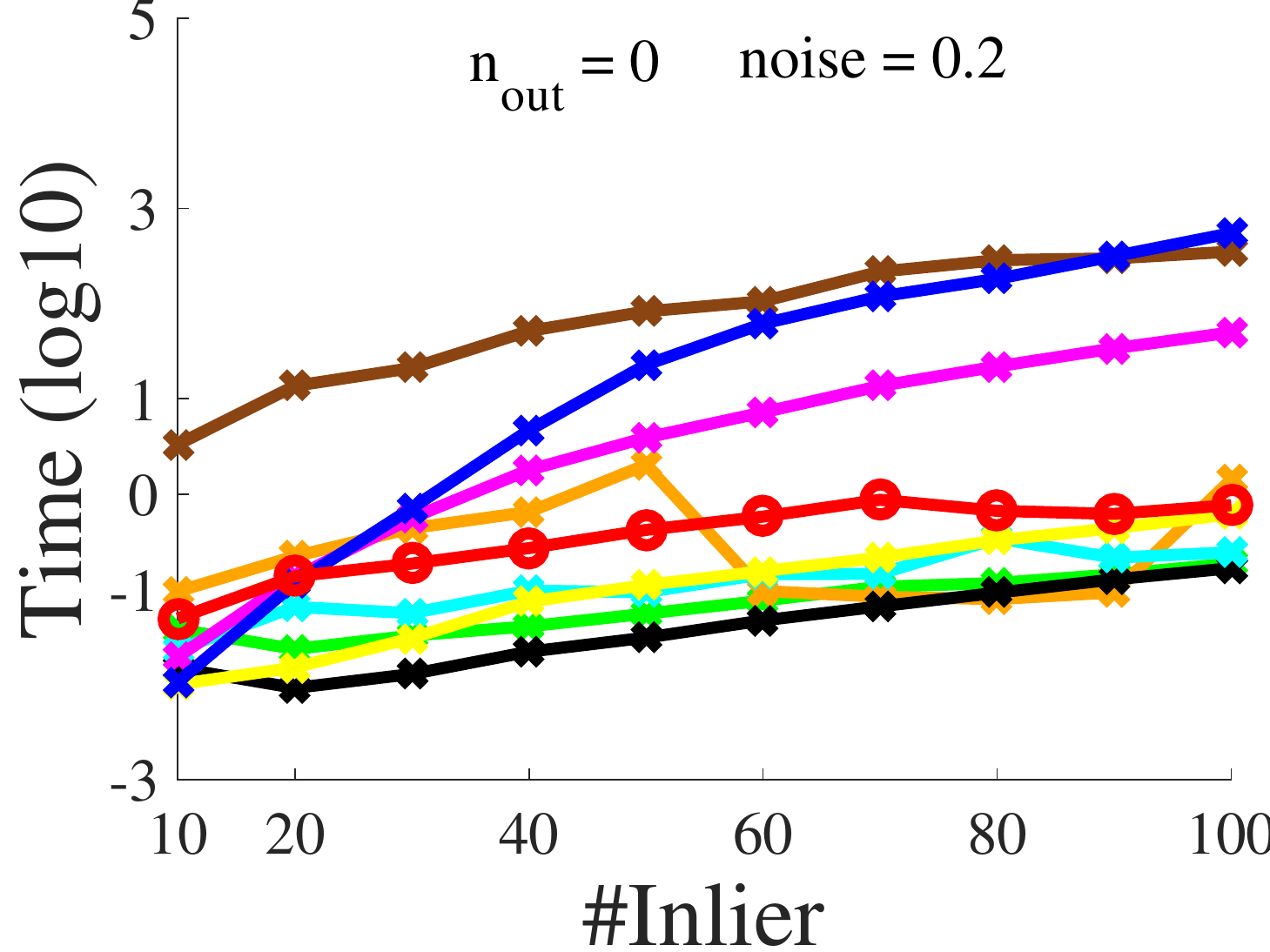}}
	\subfigure[]
	{\includegraphics[width=0.3\linewidth]{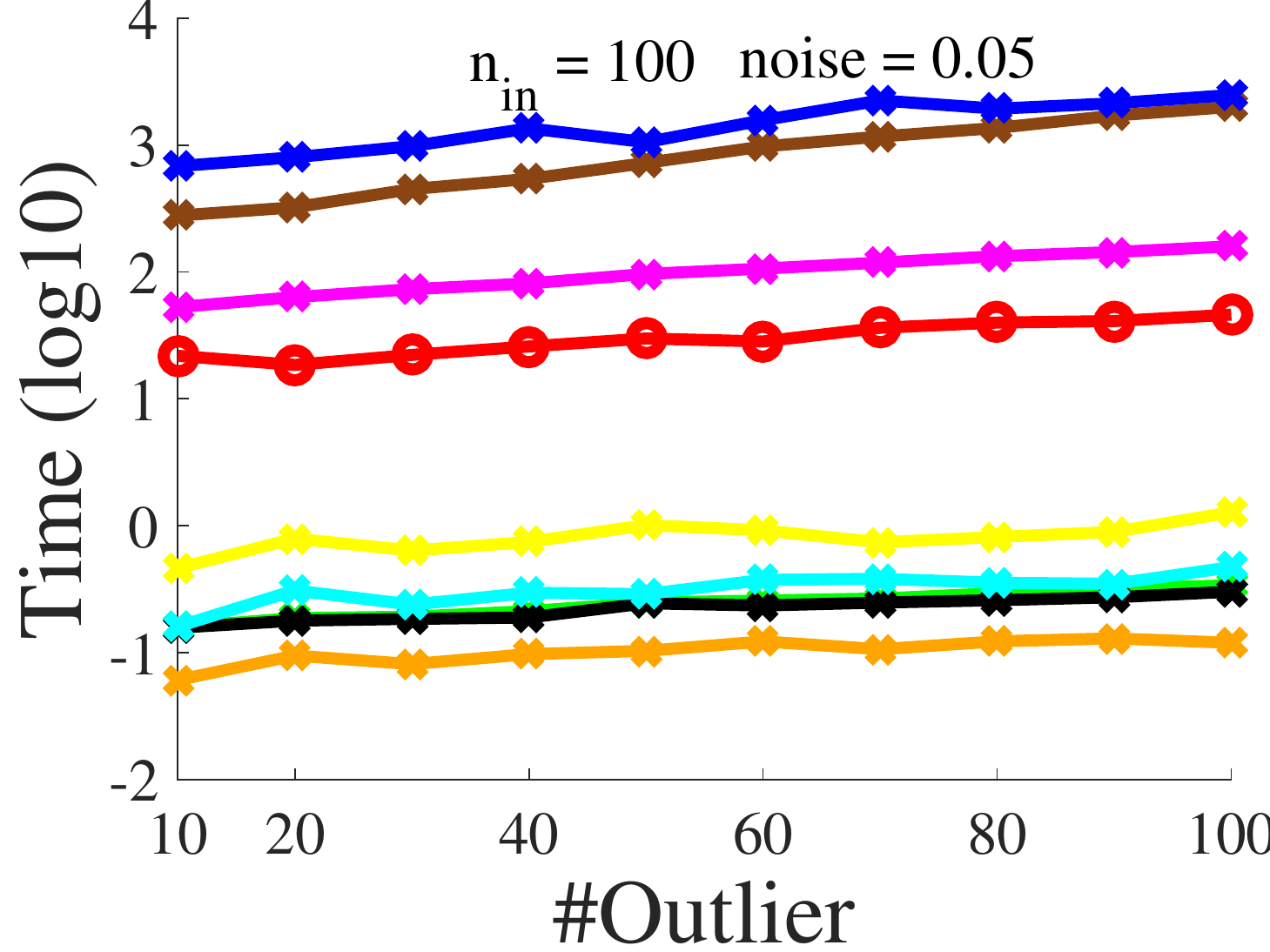}}
	\caption{Comparisons of running time and average accuracy. Graphs in (a) are complete, and those in (b) are Delaunay triangulation-connected. In (c), only ATGM uses complete graphs, while the others use Delaunay triangulation-connected graphs.}
	\label{fig_syn_time}
\end{figure}

As shown in Fig.\ref{fig_syn_time} (a) and (b), where graphs are either complete or connected by Delaunay triangulation, our method takes an intermediate running time and achieves the highest average accuracy. As shown in Fig.\ref{fig_syn_time} (c), even though ATGM handles more edges than the other methods, it takes an acceptable time with the highest accuracy. Compared with GA, SM, PM, SMAC, and IPFP-S, which run faster, ATGM can achieve higher average accuracy. To match complete graphs, the methods RRWM, FGM, MPM can achieve competitive accuracy with ATGM. However, the time consumptions of them rapidly increase and becomes larger than that of ours.

\begin{table}[htb!]
	\centering
	\caption{Average accuracy and running time of ATGM on synthetic data with varying inliers $n_{in}$, deformation noise $\sigma$ and outliers $n_{out}$.}
	\begin{minipage}{0.49\linewidth}
		\centering
		\scriptsize
		\begin{tabular}{cc|ccccc}				
			\toprule[1.0pt]	 
			\multicolumn{1}{c}{\#Inlier}&{Noise ($\sigma$)}&0.02 & 0.04  & 0.06 & 0.08 & 0.10   \\
			\hline
			\multirow{2}{20pt}{100}
			&time (s) &0.22&0.51&0.74&0.78&1.01 \\
			&acc. (\%) &99.10 &94.15 &89.75 &84.2 &73.9 \\
			\hline		
			\multirow{2}{20pt}{300}
			&time (s) &3.34&5.43&6.72&7.73&8.02 \\
			&acc. (\%) &96.87 &88.33 &74.37 &60.13 &51.33 \\
			\hline			
			\multirow{2}{20pt}{500}
			&time (s)&23.33&32.47&33.12&33.81&35.24 \\
			&acc. (\%)&94.20 &79.96 &62.32 &48.54 &38.72 \\
			\hline 
			\multirow{2}{20pt}{1000}
			&time (s)&147.15&150.92&156.71&156.99&159.26 \\
			&acc. (\%)&89.43 &66.34 &45.23 &33.47 &25.27 \\
			\bottomrule[1.0pt]	 			
		\end{tabular}		
	\end{minipage}							
	\begin{minipage}{0.49\linewidth}
		\centering
		\scriptsize
		\begin{tabular}{cc|ccccc}				
			\toprule[1.0pt]	 
			\multicolumn{1}{c}{\#Inlier}&{\#Outlier} &0.2 &0.4  & 0.6 & 0.8 & 1.0   \\
			\hline	
			\multirow{2}{20pt}{100}
			&time (s)&1.90&3.11&3.81&4.62&5.51\\
			&acc. (\%)&99.90 &99.80&99.90 &99.80 &99.60 \\
			\hline				
			\multirow{2}{20pt}{300}
			&time (s)&17.02&22.70&42.92&47.70&55.13\\
			&acc. (\%)&100.00 &99.80&99.67 &99.70 &99.53 \\
			\hline
			\multirow{2}{20pt}{500}
			&time (s)&107.24&123.42&146.99&187.84&185.54\\
			&acc. (\%)&99.86 &99.88 &99.64 &98.24 &81.30 \\
			\hline
			\multirow{2}{20pt}{1000}
			&time (s)&563.83&645.11&758.73&882.18&1070.26\\
			&acc. (\%)&99.84 &98.95 &88.44 &78.17 &71.33 \\
			\bottomrule[1.0pt]	 
		\end{tabular}	
	\end{minipage}
	\label{table_largescale}
\end{table}

{\bf Large-scale graph matching.}
To test the efficiency of our method when applied to large-scale graphs, we carried out more challenging experiments by setting the number of inliers as $n_{in}=100,300,500,1000$ with deformation noise and outliers. The number of outliers was set to $20\%,40\%,...,100\%$ of the number of inliers.

As reported in Tab.\ref{table_largescale}, ATGM is very robust to outliers and less robust to strong noise with larger graphs. Since the compared methods need to store affinity matrices with size of approximately $n_{in}^2(n_{in}+n_{out})^2$, applying these methods to large-scale graphs with hundreds or thousands of nodes is infeasible.

\subsection{Results on real-world datasets}
We also perform comparative evaluations on real-world datasets, including the CMU House sequence\footnote{\url{http://vasc.ri.cmu.edu//idb/html/motion/house/index.html}} and the PASCAL Cars and Motorbikes pairs~\cite{[2012-Leordeanu-ijcv]}, which are commonly used to evaluate graph matching algorithms.

\begin{figure}[htb!]
	\centering 
	\subfigure[20-vs-30 (ATGM:~20/20)] 
	{\includegraphics[height=0.11\linewidth,width=0.32\textwidth]{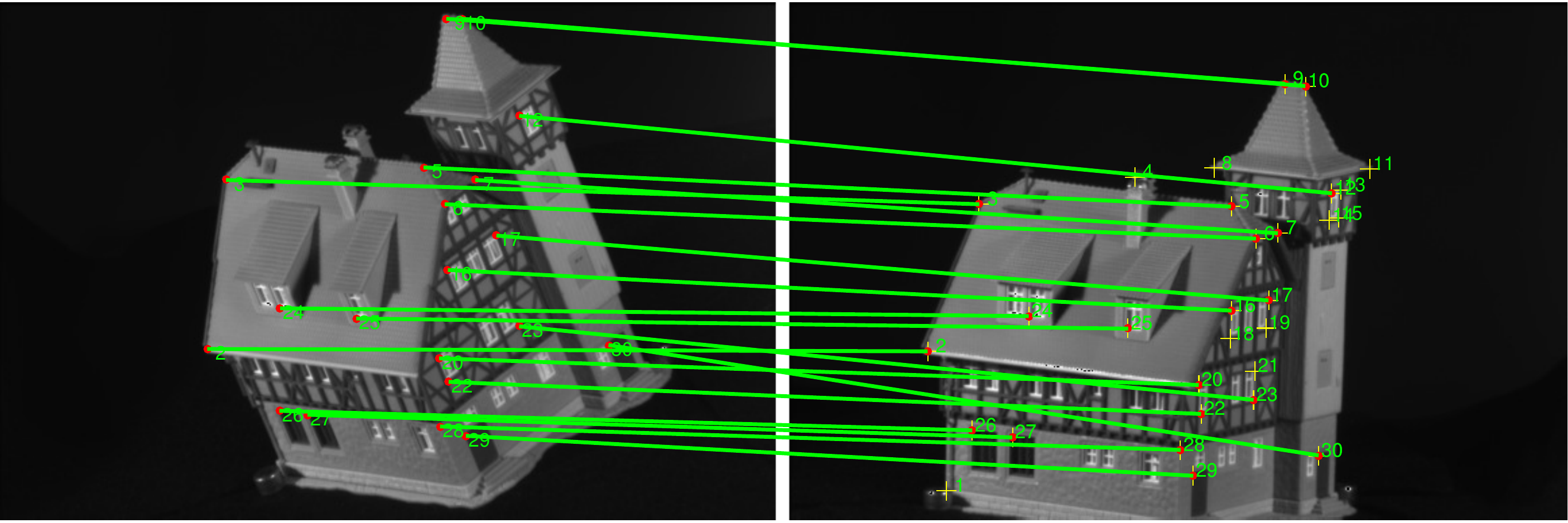}}
	\subfigure[28-vs-48 (ATGM:~28/28)]  	    
	{\includegraphics[height=0.11\linewidth,width=0.32\textwidth]{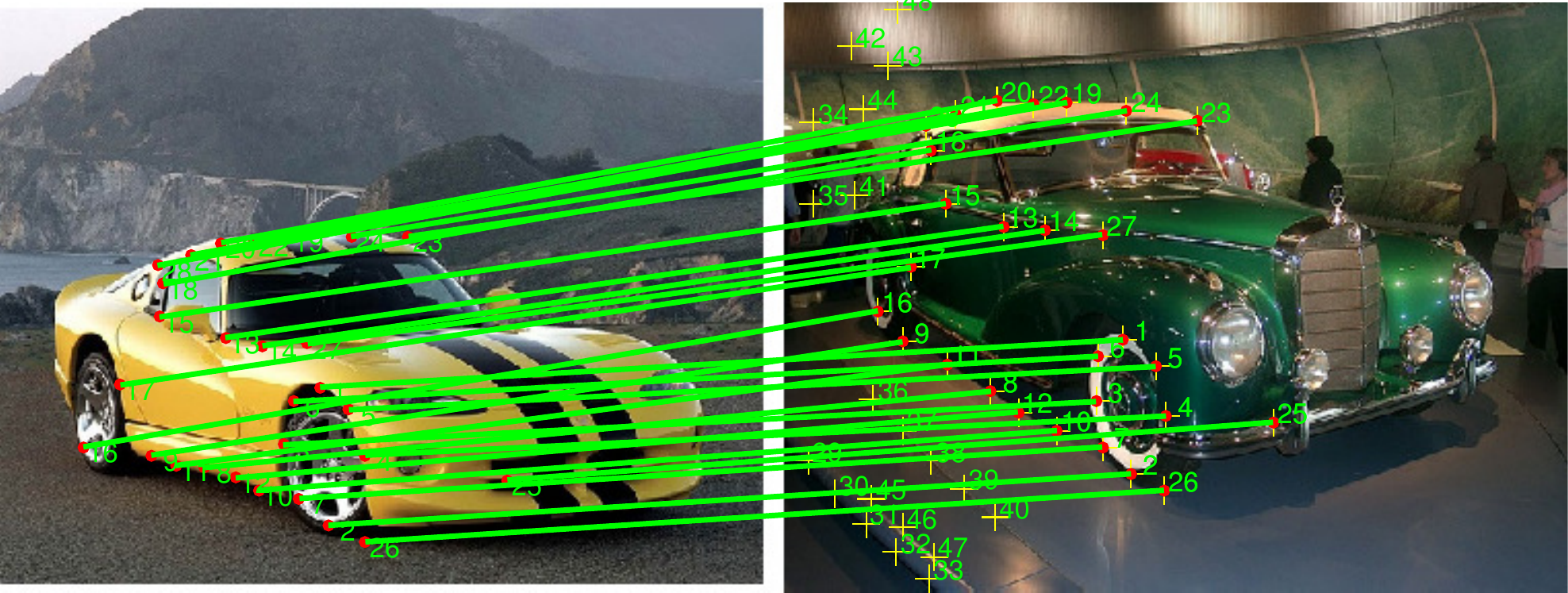}}
	\subfigure[46-vs-86 (ATGM:~44/46)]  	    
	{\includegraphics[height=0.11\linewidth,width=0.32\textwidth]{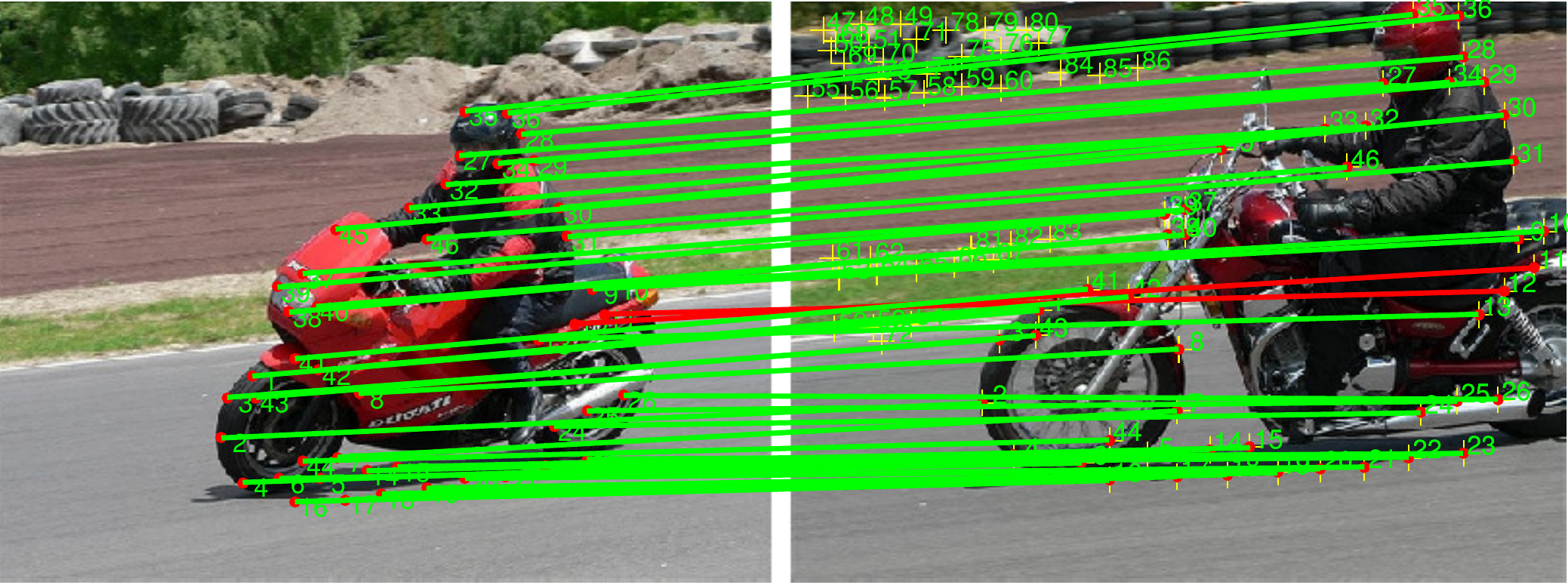}}
	\caption{Examples of matching unequal-size graphs using ATGM on real-world datasets. The red dots are inliers in $\mathcal{G}_X$, and yellow plus signs are both inliers and outliers in $\mathcal{G}_Y$. The lines in green are correct matches, while those in red are incorrect.}
\end{figure}

\begin{figure}[htb!]
\centering
		\includegraphics[width=0.8\linewidth]{./Images/fig_henglan.pdf}\\
		\subfigure[House:30-vs-30]
		{\includegraphics[width=0.32\linewidth]{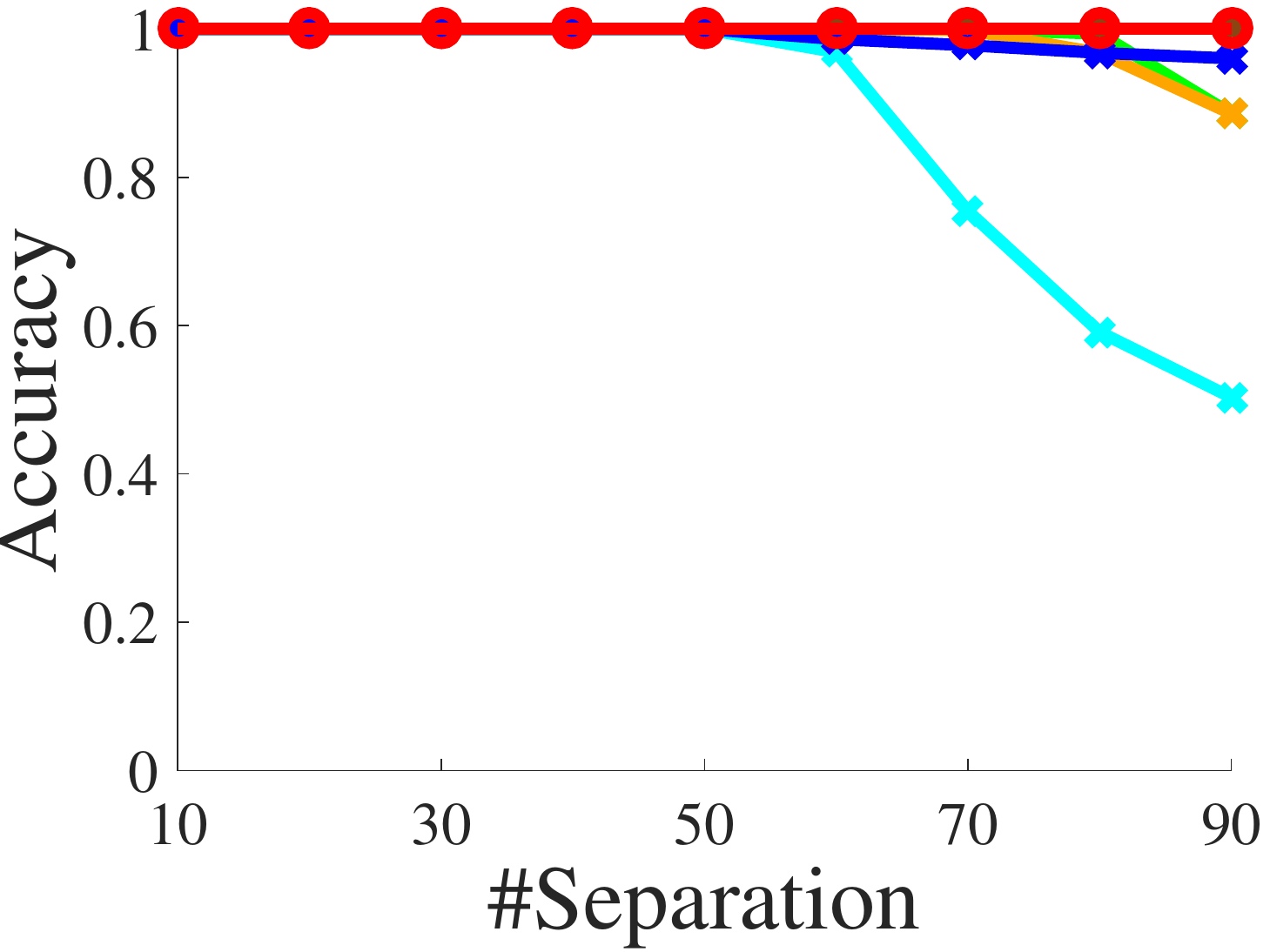}}
		\subfigure[House: 25-vs-30]
		{\includegraphics[width=0.32\linewidth]{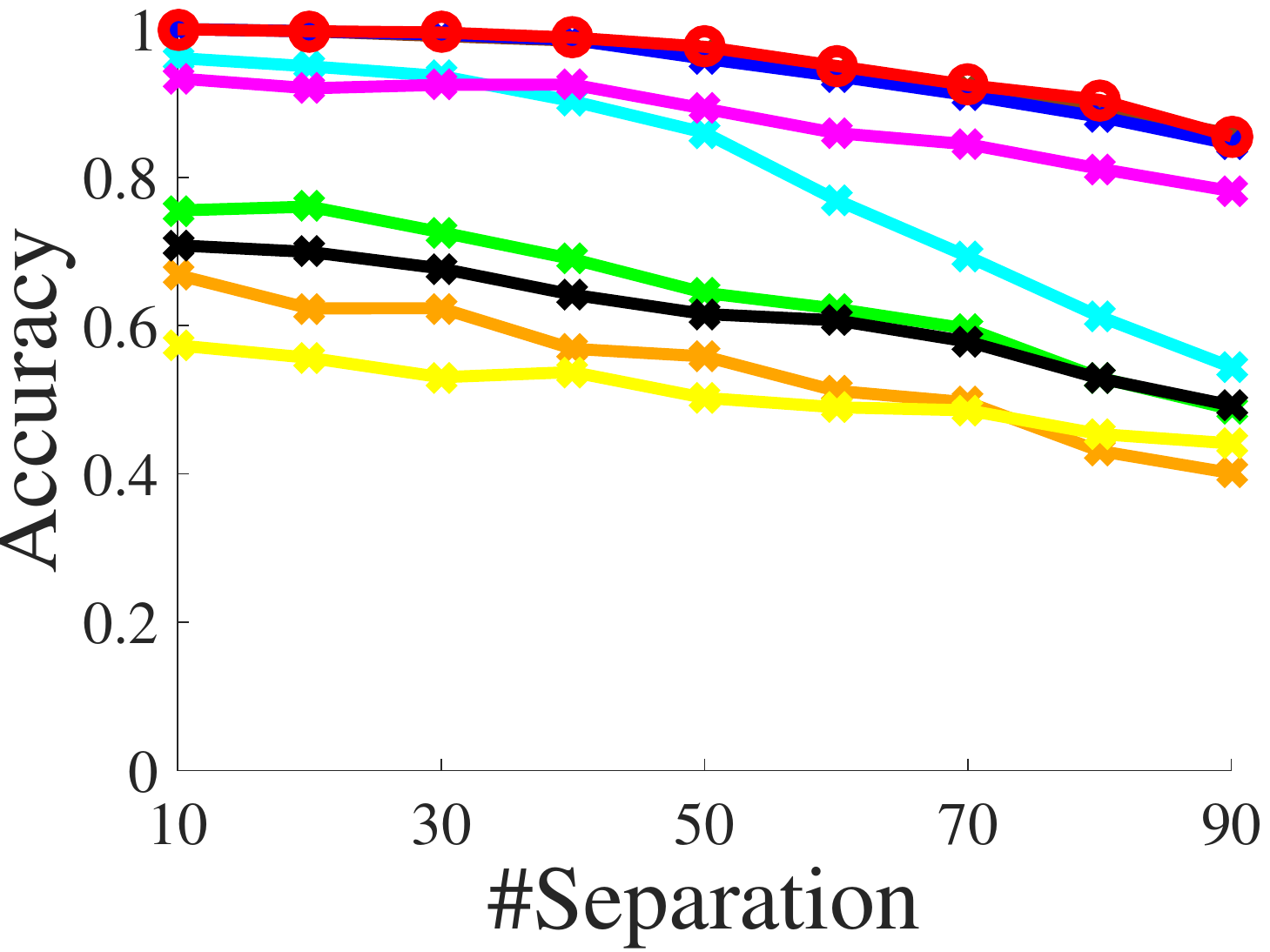}}
		\subfigure[House: 20-vs-30]
		{\includegraphics[width=0.32\linewidth]{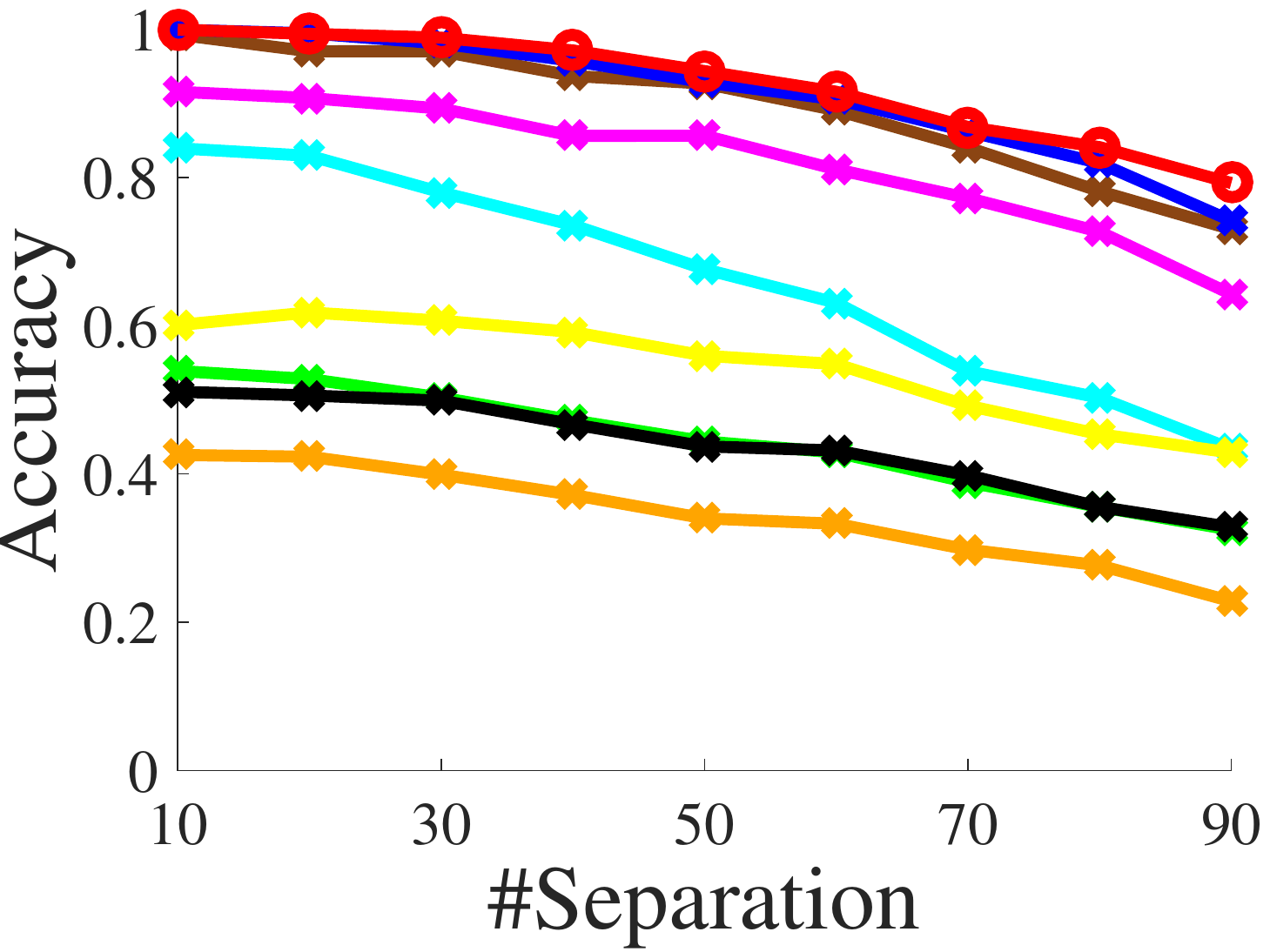}}	
	\caption{Comparison of average accuracy on the House sequence in both equal-size and unequal-size cases.}\label{fig:CMU-house}
\end{figure}

\begin{figure}[htb!]
	\centering
	{\includegraphics[width=0.9\linewidth]{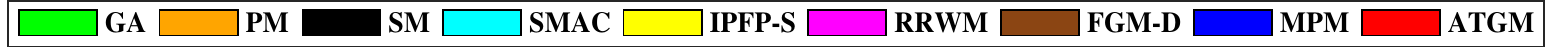}}\\
	{\includegraphics[width=0.48\linewidth]{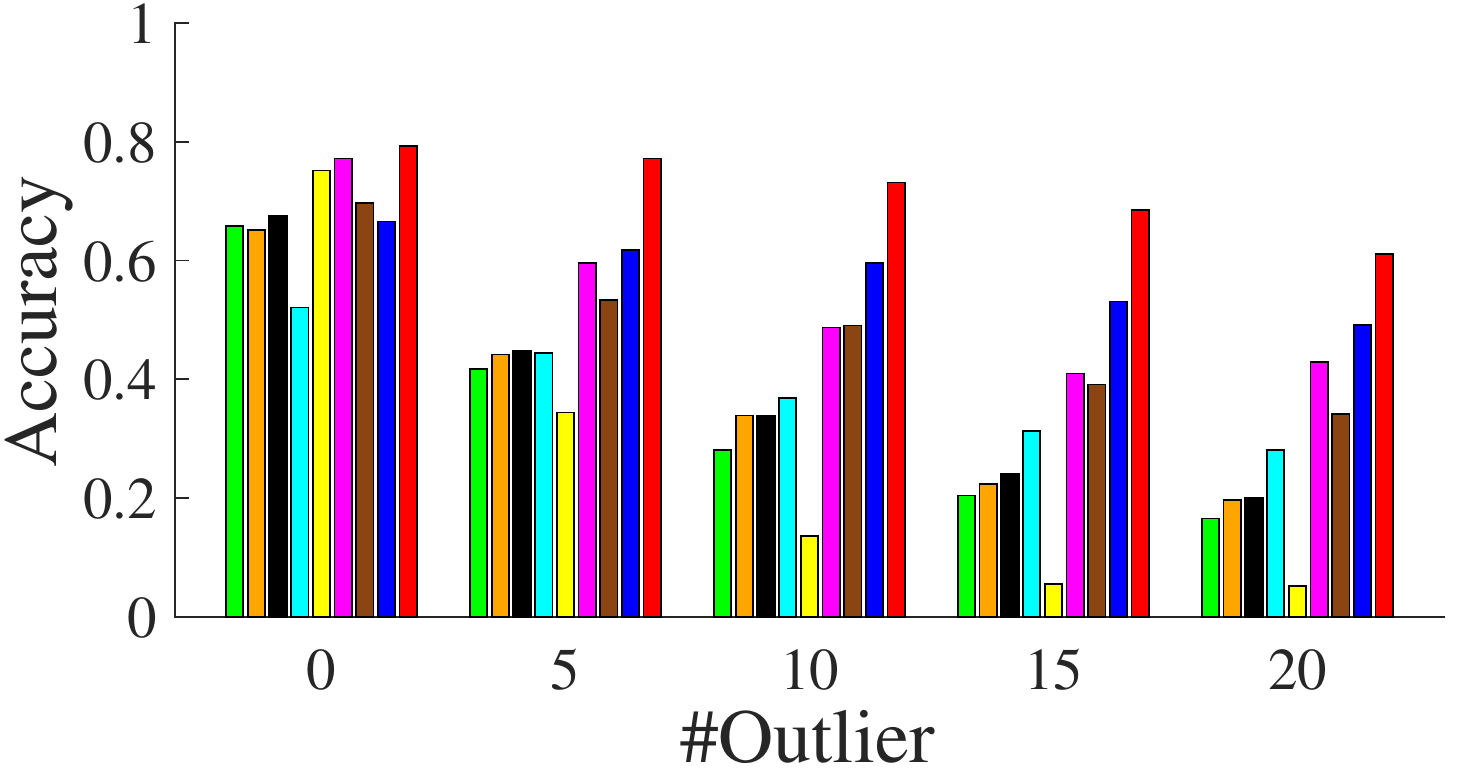}}
	{\includegraphics[width=0.48\linewidth]{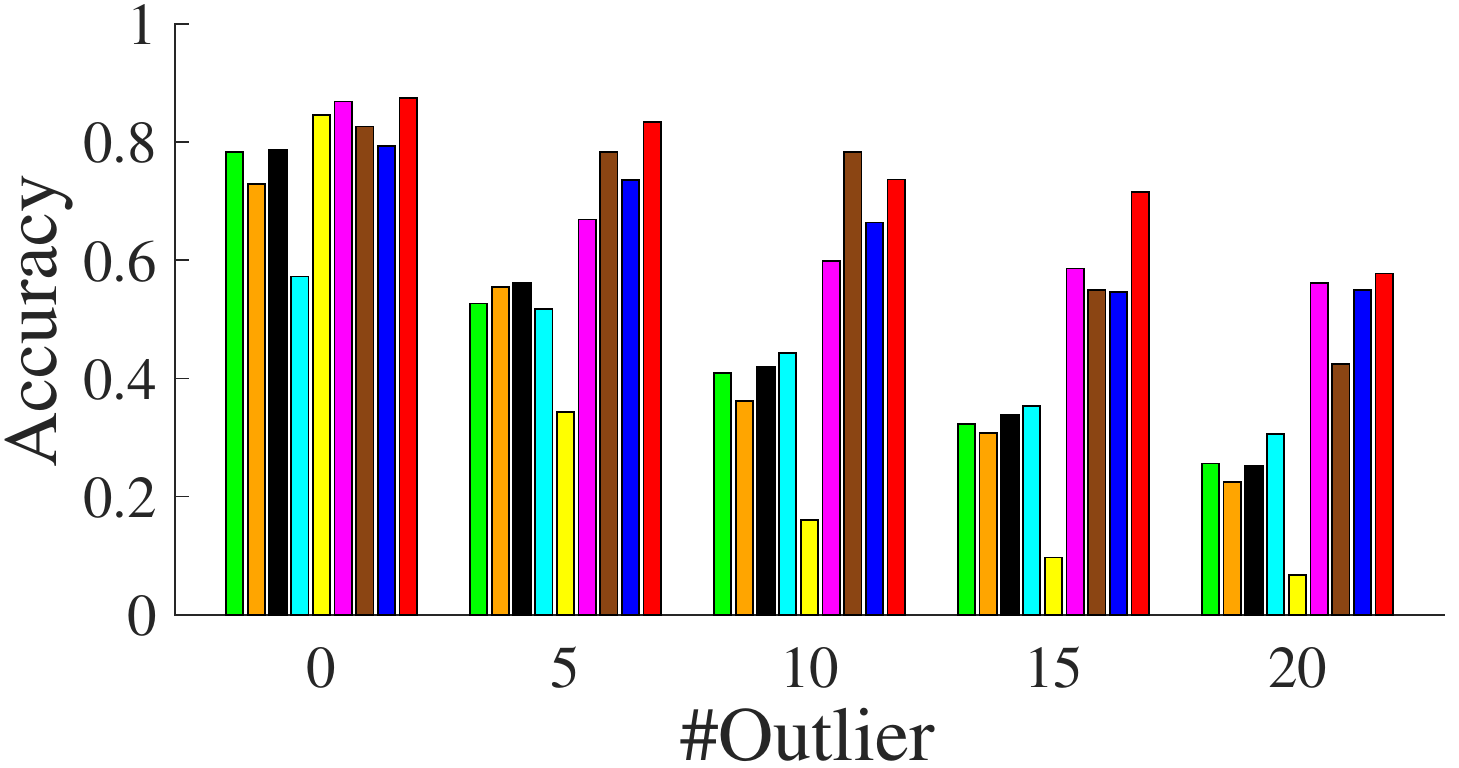}}
	\caption{Comparison on cars (left) and motorbikes (right) image pairs with outliers.}
	\label{fig:car-motor}
\end{figure}

The CMU House sequence consists of 111 frames of a synthetic house. Each image contains 30 feature points that are manually marked with known correspondences. In this experiment, we matched all the image pairs separated by 10, 20,.., 90 frames. The unequal-size cases are set as 20-vs-30 and 25-vs-30. For the compared methods, we set the edge-affinity to $\mathbf{W}_{i_1j_1,i_2j_2}=\text{exp}(-\frac{(||X_{i_1i_2}||-||Y_{j_1j_2}||)^2}{2500})$ as the same as ~\cite{[2016-Zhou-pami]}.

The PASCAL dataset for graph matching consists of 30 pairs of car images and 20 pairs of motorbike images. Each pair contains both inliers (approximately 30--60 feature points) with groundtruth labels and randomly marked outliers. In the unequal-size matching case, we added 5, 10, 15, 20 outliers to $\mathcal{G}_Y$. For the compared methods, we set the edge affinity matrix as Eq.\eqref{equationedge} which was used in~\cite{[2016-Zhou-pami]}.

{\bf Average accuracy} 
For the CMU House sequence, as shown in Fig.\ref{fig:CMU-house}, our method achieves a higher accuracy in both equal-size and unequal-size cases. Meanwhile, our method outperforms all the compared methods on the PASCAL datasets because our method can remove the outliers automatically. The results are shown in Fig.\ref{fig:car-motor}.

{\bf Effect of objective functions} As we discussed in Sec.\ref{method}, the objective function $G_{\bar{X}Y}$ has effects on both the sparsity and matching accuracy. First, to evaluate the sparsity of $\mathbf{P}\in [0,1]^{m\times n}$, we define an index $S_r(\mathbf{P})= \frac{\sum_{i}\mathbb{I}_{(\mathbf{P}_{ij}\ge r)} }{m}$ where $\mathbb{I}$ is the indicator function. We evaluated $S_r(\mathbf{P})$ on the House sequence with $r=0.9$. As shown in Tab.\ref{tab_acc_sparse}, the optimal representation map $\mathbf{P}^*$ of $G_{\bar{X}Y}$ is (nearly) binary in all cases. Then, we evaluated the average accuracy in two cases: (1) minimizing $F_{XY}$ only and (2) applying $G_{\bar{X}Y}$ after $F_{XY}$ is minimized. As shown in Tab.\ref{tab_acc_sparse}, the average accuracy is highly improved especially in unequal-size cases due to $G_{\bar{X}Y}$. This results shows that $G_{\bar{X}Y}$ can enhance the sparsity of the assignment matrix and reduce the node shifting.

\begin{table}[htb!]
	\centering
	\footnotesize
	\caption{Effect of objective functions $F_{XY}$ and $G_{\bar{X}Y}$ on both the sparsity of the assignment matrix $\mathbf{P}^*$ and the average matching accuracy of the house sequence dataset.}
	\begin{tabular}{c|c|ccccccccc}
		\toprule[0.8pt]		
		\multirow{1}{30pt}{Size}&\#Separation &10& 20 & 30 & 40&50 & 60& 70 & 80& 90 \\	\hline	
		\multirow{3}{30pt}{m=20\\n=30}
		& Sparsity&98.18&97.98&97.59&96.11&95.15&89.63&90.79&79.58&80.90\\
		\cline{2-11}
		& acc. (F) &59.60&58.89&57.78&56.18&55.38&54.05&53.52&51.90&51.39\\
		&acc. (F\&G)&\textbf{98.25}&\textbf{97.86}&\textbf{96.84}&\textbf{93.97}&\textbf{92.11}&\textbf{88.37}&\textbf{85.66}&\textbf{79.37}&\textbf{77.67}\\
		\hline
		\multirow{3}{30pt}{m=25\\n=30}
		&Sparsity&99.92&100.00&100.00&99.72&99.42&98.35&98.42&96.63&92.43\\
		\cline{2-11}
		&acc. (F) &81.25&80.24&78.15&76.56&75.80&74.93&73.25&71.05&68.92\\
		&acc. (F\&G)&\textbf{99.92}&\textbf{99.71}&\textbf{99.42}&\textbf{98.66}&\textbf{97.70}&\textbf{96.05}&\textbf{94.63}&\textbf{91.63}&\textbf{89.08}\\
		\hline
		\multirow{3}{30pt}{m=30\\n=30}
		& Sparsity&100.00&100.00&100.00&100.00&100.00&100.00&100.00&100.00&100.00\\
		\cline{2-11}
		& acc. (F) &100.00&100.00&100.00&100.00&100.00&100.00&100.00&100.00&99.68\\
		&acc. (F\&G)&\textbf{100.00}&\textbf{100.00}&\textbf{100.00}&\textbf{100.00}&\textbf{100.00}&\textbf{100.00}&\textbf{100.00}&\textbf{100.00}&\textbf{100.00}\\
		\bottomrule[0.8pt]
	\end{tabular}
	\label{tab_acc_sparse}
\end{table}

\begin{table}[ht!]
	\centering
	\scriptsize
	\caption{Effectiveness of outlier removal strategy. This strategy improves the average matching accuracy by more than ${\textbf{10}\%}$ for almost all the methods.}
	\begin{tabular}{cc|ccccccccc}
		\toprule[0.8pt]		
		\multirow{1}{20pt}{Data}&Out.Re. & GA~\cite{[1996-Gold]}& PM~\cite{[2008-Zass-cvpr]} & SM~\cite{[2005-Leordeanu]} & SMAC~\cite{[2006-Cour-nips]}&IPFP-S~\cite{[2009-Leordeanu-nips]} & RRWM~\cite{[2010-Cho-eccv]} & FGM-D~\cite{[2016-Zhou-pami]} & MPM~\cite{[2014-Cho-cvpr]}& {\bf ATGM} \\
		\hline			
		\multirow{2}{20pt}{Cars}
		& w/o&34.50&37.04&38.04&38.53&26.74&53.84&49.05&58.02&-\\
		& w/&\textbf{61.93}&\textbf{60.71}&\textbf{63.55}&\textbf{49.54}&\textbf{65.55}&\textbf{70.37}&\textbf{70.62}&\textbf{63.44}&\textbf{71.83}\\
		\hline
		\multirow{2}{15pt}{Motor.}
		& w/o& 45.97& 43.56&47.13& 43.84&34.90& 65.64&67.31& 65.73& -\\
		& w/&\textbf{ 66.53}&\textbf{61.91}&\textbf{67.43}&\textbf{52.06} &\textbf{75.80}& \textbf{72.61}&\textbf{76.76}&\textbf{69.46}&\textbf{74.75}\\		
		\bottomrule[0.8pt]
	\end{tabular}
	\label{table_re}
\end{table}

{\bf Effectiveness on outlier removal} Finally, our proposed outlier removal strategy is not restricted to our approach. It can be applied to any other method. To evaluate the generality of our outlier removal strategy, we applied it as a pre-processing step, and then executed the other methods with the pre-processed input. As shown in Tab.\ref{table_re}, the average accuracy of all the methods is improvement greatly, and almost all the methods improve their performance by more than $10\%$.

\section{Conclusions}\label{section6}
In this paper, we presented a new approach from a functional representation perspective for the graph matching problem by redefining the assignment matrix as a linear representation map. Our approach reduces both the space and time complexity significantly. Thus, our method is suitable for matching complete graphs with hundreds and thousands of nodes. In addition to the transformation map, we presented a domain adaptation-based method for outlier removal that improves the performance of all methods. In future work, we plan to study graph matching on more general manifolds (or metric spaces) and hyper-graph matching with lower computational complexity.

\section{Acknowledgement}
This research is supported by projects of National Natural Science Foundation of China (NSFC) under the contracts No.61771350 and No.41501462. 
%
%

%
%
 \bibliographystyle{splncs04}
 \bibliography{egbib_eccv}
%
%
%
%
%


\end{document}